%% file: main_arxiv.tex
\documentclass{article}
\input{commons}

\pdfoutput=1
\usepackage{graphicx}
\usepackage{subfig}
\usepackage{booktabs} 
\usepackage{xcolor}
\usepackage{etoolbox}
\usepackage{anyfontsize}
\usepackage{wrapfig}
\usepackage{caption, physics}

\usepackage{tikz}
\usepackage{pgfplots}
\usetikzlibrary{spy,calc}

\usepackage{float}
\usepackage{algorithmic, algorithm}

\usepackage[colorlinks=true,linkcolor=magenta,citecolor=green!80!black]{hyperref}

\makeatletter
\newcommand{\org@footnotemark}{}
\let\org@footnotemark\@footnotemark
\def\@footnotemark{%
\begingroup
\hypersetup{linkcolor=blue!70}%
\org@footnotemark
\endgroup
}
\makeatother

\usepackage[nonatbib, final]{diffeqml}
\usepackage[round]{natbib}
\usepackage[many]{tcolorbox}

\usepackage[toc,page,header]{appendix}
\usepackage{minitoc}
\numberwithin{equation}{section}

\title{\fontsize{17}{19}\selectfont Self--Similarity Priors:\\ Neural Collages as Differentiable Fractal Representations}
\author{\hspace{0.9cm}Michael Poli\thanks{Equal contribution authors.}\\
	\hspace{0.9cm}{\normalsize Stanford University}\\
	\hspace{0.9cm}\fontsize{9}{10}\selectfont{\texttt{poli@stanford.edu}}
    \And
    \hspace{1.25cm}Winnie Xu$^{\color{blue!70}*}$\\
	\hspace{1.25cm}{\normalsize University of Toronto}\\
	\hspace{1.25cm}\fontsize{9}{10}\selectfont{\texttt{winniexu@cs.toronto.edu}}
    \And\normalsize
    \hspace{0.35cm}Stefano Massaroli$^{\color{blue!70}*}$\\
	\hspace{0.35cm}\normalsize University of Tokyo\\
	\hspace{0.35cm}\fontsize{9}{10}\selectfont$\tt massaroli\texttt{@}robot.t.u\text{-}tokyo.ac.jp$
	\AND\normalsize\hspace{-0.65cm}
    Chenlin Meng\\\hspace{-0.65cm}
    \normalsize	Stanford University\\
	\And \normalsize\hspace{0.3cm}
    Kuno Kim\\\hspace{0.3cm}
    \normalsize	Stanford University\\
	\And \normalsize\hspace{0.3cm}
	Stefano Ermon\\\hspace{0.3cm}
    \normalsize	Stanford University
}

\begin{document}

\maketitle

\begin{abstract}
Many patterns in nature exhibit \textit{self--similarity}: they can be compactly described via self--referential transformations. Said patterns commonly appear in natural and artificial objects, such as molecules, shorelines, galaxies and even images. In this work, we investigate the role of learning in the automated discovery of self-similarity and in its utilization for downstream tasks. To this end, we design a novel class of implicit operators, Neural $\COs$, which (1) represent data as the parameters of a self--referential, structured transformation, and (2) employ hypernetworks to amortize the cost of finding these parameters to a single forward pass. We investigate how to leverage the representations produced by Neural $\COs$ in various tasks, including data compression and generation. Neural $\CO$ image compressors are orders of magnitude faster than other self--similarity--based algorithms during encoding and offer compression rates competitive with implicit methods. Finally, we showcase applications of Neural $\COs$ for fractal art and as deep generative models.

\end{abstract}

\input{sections_arxiv/01_intro}

\input{sections_arxiv/02_preliminaries}
\input{sections_arxiv/03_collage_ops}

\input{sections_arxiv/05_experiments}
\input{sections_arxiv/06_related_conclusions}
\bibliographystyle{abbrvnat}
\bibliography{biblio.bib}

\newpage

\begin{center}
    \huge{\bf{Self--Similarity Priors} \\
    \emph{Supplementary Material}}
\end{center}
\vspace*{3mm}

\appendix
\addcontentsline{toc}{section}{}
\part{}
\parttoc

\input{appendix_arxiv/B_problem_setting}
\input{appendix_arxiv/C_additional}

\input{appendix_arxiv/D_empirics}
\input{appendix_arxiv/E_genmo}
\end{document}

%% file: commons.tex
\usepackage{minitoc}
\usepackage{amsmath,amsthm, mathtools}
\usepackage{amssymb}
\usepackage{xcolor,colortbl}
\usepackage{enumitem}
\usepackage[toc,page,header]{appendix}
\usepackage{minitoc}
\usepackage{pst-node}

\setlist{leftmargin=2.5mm}
\usepackage{tikz}
\usetikzlibrary{shapes.geometric}
\usetikzlibrary{spy}

\usepackage{thmtools, thm-restate}
\declaretheorem{theorem}

\newtheorem{lemma}{Lemma}
\newtheorem{cor}{Corollary}
\newtheorem{defn}{Definition}
\newtheorem{remark}{Remark}

\newtheorem{example}[cor]{Example}
\newcommand{\x}{\times}

\newcommand{\cE}{\mathcal{E}}
\newcommand{\cF}{\mathcal{F}}
\newcommand{\cH}{\mathcal{H}}

\newcommand{\cX}{\mathcal{X}}

\newcommand{\sD}{\mathsf{D}}

\newcommand{\sI}{\mathsf{I}}

\newcommand{\sR}{\mathsf{R}}
\newcommand{\sS}{\mathsf{S}}

\newcommand{\sU}{\mathsf{U}}

\DeclareMathAlphabet{\nummathbb}{U}{BOONDOX-ds}{m}{n}

\newcommand{\1}{\nummathbb{1}}

\newcommand{\bA}{\mathbb{A}}

\newcommand{\bD}{\mathbb{D}}
\newcommand{\bE}{\mathbb{E}}

\newcommand{\Id}{\mathbb{I}}

\newcommand{\bN}{\mathbb{N}}

\newcommand{\R}{\mathbb{R}}
\newcommand{\bS}{\mathbb{S}}

\newcommand{\bU}{\mathbb{U}}
\newcommand{\bW}{\mathbb{W}}
\newcommand{\bX}{\mathbb{X}}
\newcommand{\bY}{\mathbb{Y}}
\newcommand{\bZ}{\mathbb{Z}}

\newcommand{\CO}{{\tt Collage} }
\newcommand{\COs}{{\tt Collages}}

\DeclareMathOperator*{\dom}{\mathsf{dom}}

\definecolor{olive}{rgb}{0.6, 0.6, 0.2}
\definecolor{sand}{rgb}{0.8666666666666667, 0.8, 0.4666666666666667}
\definecolor{wine}{rgb}{0.5333333333333333, 0.13333333333333333, 0.3333333333333333}
\definecolor{deblue}{RGB}{11,132,147}
\definecolor{ocra}{RGB}{204, 119, 34}
\definecolor{depurple}{RGB}{131, 102, 135}
\definecolor{degrey}{RGB}{186, 172, 172}
\newcommand{\fcircle}[2][red,fill=red]{\tikz[baseline=-0.5ex]\draw[#1,radius=#2] (0,0.03) circle ;}

\makeatletter
\def\@footnotecolor{red}
\define@key{Hyp}{footnotecolor}{%
 \HyColor@HyperrefColor{#1}\@footnotecolor%
}
\def\@footnotemark{%
    \leavevmode
    \ifhmode\edef\@x@sf{\the\spacefactor}\nobreak\fi
    \stepcounter{Hfootnote}%
    \global\let\Hy@saved@currentHref\@currentHref
    \hyper@makecurrent{Hfootnote}%
    \global\let\Hy@footnote@currentHref\@currentHref
    \global\let\@currentHref\Hy@saved@currentHref
    \hyper@linkstart{footnote}{\Hy@footnote@currentHref}%
    \@makefnmark
    \hyper@linkend
    \ifhmode\spacefactor\@x@sf\fi
    \relax
  }%
\makeatother

%% file: sections_arxiv/01_intro.tex
\section{Introduction}
\begin{center}
    \small \textit{Given a specified image, can one come up with a dynamical system with it as its attractor? \citep{welstead1999fractal}}
\end{center}
Scientific fields are underpinned by a search for structure. Geometry, sparsity and invariances, when appropriately introduced in a mechanistic model, allow us to concisely describe phenomena. To this end, machine learning has been introduced as a means to fix partial priors in a model, and discover the rest through data \citep{rackauckas2020universal,dao2020kaleidoscope,bronstein2021geometric}. In general, the notion of structure is also essential for compression: through a suitable choice of language, one can explain phenomena in fewer symbols, yielding shorter representations of the observables \citep{tishby2000information,lee2007efficient}.

\begin{wrapfigure}[16]{r}{0.51\textwidth}
    \centering
    \includegraphics[width=0.99\linewidth]{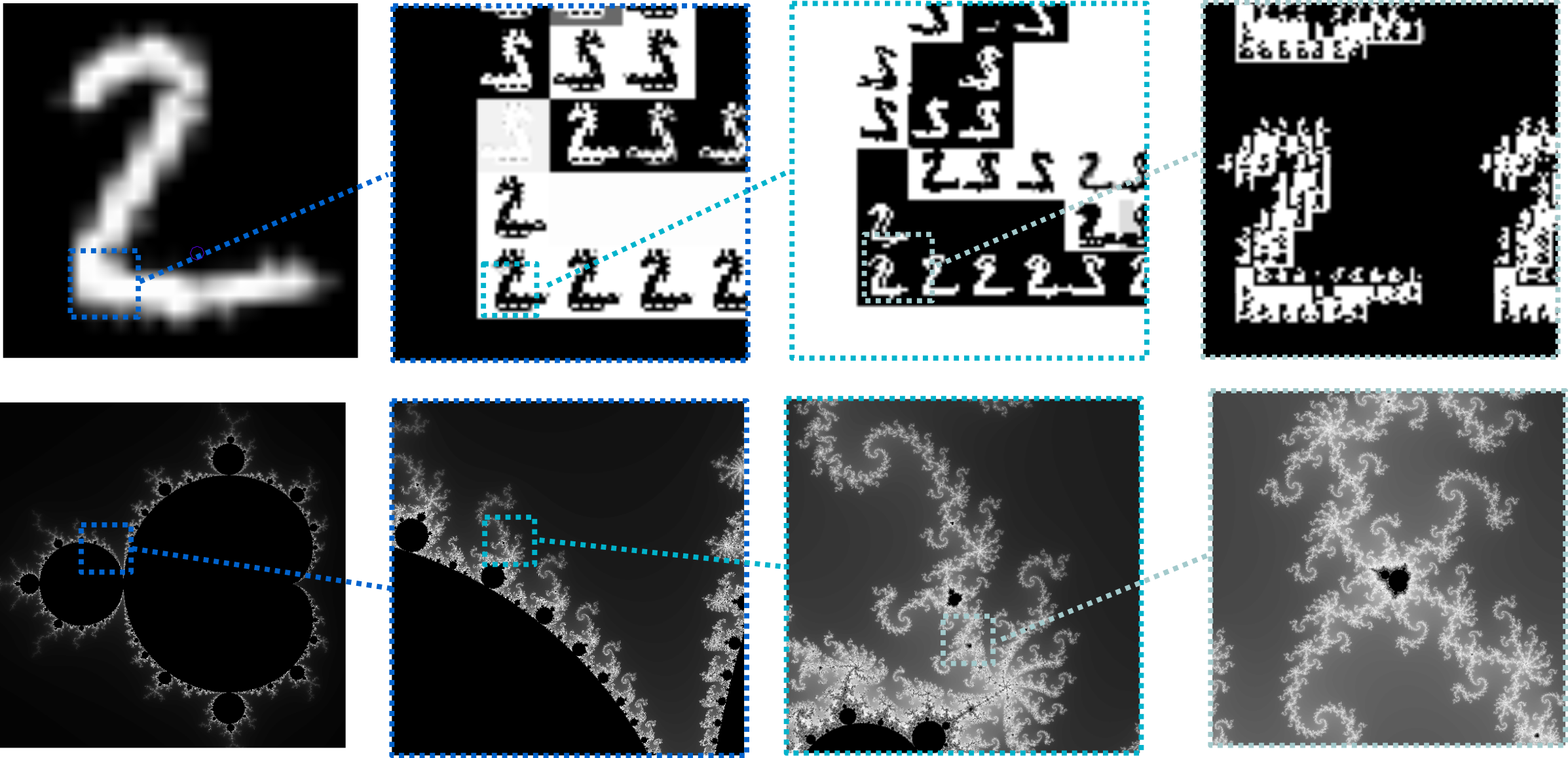}
    \vspace{-1mm}
    \caption{\small \textbf{[Top]} MNIST digit \textit{fractalized} via a $\CO$. The image is represented as the coefficients of a $\CO$, and decoded as its attractor. Magnification is done by decoding at higher resolutions. \textbf{[Bottom]} The Mandelbrot set \citep{mandelbrot1980fractal}, an example of a fractal displaying self--similarity.}
    \vspace{-5mm}
    \label{fig:1mandelbrot}
\end{wrapfigure}

Objects exhibiting \textit{self--similarity} structure are composed of patterns that appear similar to themselves at multiple scales, as shown in Figure \ref{fig:1mandelbrot}. This type of structure frequently appears in nature, at different degrees: shorelines, molecules, plants, turbulent flows and basins of attraction of dynamical systems all display elements of self--similarity \citep{mandelbrot1982fractal,song2005self,vulpiani2009chaos,barnsley2014fractals}.
In this work, we explore the role of learning in the automatic discovery of \textit{self--similarity} structure in data, and how it can serve as an inductive bias in machine learning models.

The mathematical embodiment of this idea is found in fractal patterns, which often arise by characterizing limit sets of nonlinear maps e.g. iterations of complex numbers for Julia and Mandelbrot sets~\citep{julia1918memoire,mandelbrot1980fractal}. 
Fractals are scale--invariant: they can be "zoomed in" by increasing the resolution of the limit sets, and manifest arbitrarily similar patterns at different scales. Despite their apparent infinite complexity, a fractal can be uniquely and compactly described by its generating nonlinear map. 

A method to discover self--similar structure in data (not necessarily of fractal nature) can then be formalized as an optimization problem: after choosing an appropriate class of contractive, parameterized maps, one searches for 
parameters such that a given data point can be (approximately) recovered as the (unique) fixed--point of the chosen map. 
This approach, pioneered in \citep{barnsley1986solution}, paved the way for one of the most successful algorithmic applications of self--similarity, \textit{fractal image compression} \citep{jacquin1992image,jacquin1993fractal,barnsley1996fractal,welstead1999fractal,fisher2012fractal}. 
First, an encoding step carries out a search to solve the inverse problem of data to operator parameters via extensive search, or by restricting the class of operators such that a closed--form solution may be found. Then, given the parameters, the decoding step solves for the fixed--point of the operator, corresponding to a corrupted version of the original data. The quality of decoded images i.e. the \textit{loss} of the fractal compression method is directly tied with the expressivity of the class of operators considered, which are often designed to seek  self--similarities in pixel space. Yet, larger classes induce more challenging optimization problems, leading to long encoding times.

Despite unique properties, such as high compression rates in data with a high degree of self--similarity \citep{welstead1999fractal} and the ability to magnify images during decoding \citep{mitra2000technique}, fractal compression methods are rarely used in practice. The main limitations are slow encoding times\footnote{Even with a fully--parallelized implementation leveraging modern deep learning frameworks and GPUs, a single $1000\times1000$ RGB image takes minutes to encode with a standard PIFS fractal compression scheme.}, and poor scaling of decoded image quality at longer code lengths, due to heavy restrictions placed on the class of operators to keep the inverse problem solvable in closed form \citep{fisher2012fractal}.

Here, we propose a novel learning--based technique to extract and utilize self--similar representations of data. We develop Neural $\COs$, a family of differentiable, parametrized operators structured to capture self--similarity between partitions of data. The inverse problem of Neural $\COs$ is solved with a single forward pass of a hypernetwork \citep{ha2016hypernetworks} trained to generate a set of parameters as the fractal code. This amortized approach is  orders of magnitude faster than traditional search based methods.
$\CO$ operators are composable with neural network architectures and have wide applicability beyond compression, as they can be optimized end--to--end for a variety of tasks including generative modeling. 

In data compression, Neural $\COs$ preserve advantages of fractal compression methods, with up to $10\times$ (accounting for training time) and $100\times$ (at test time) speedups during encoding. Further, we investigate deep generative models based on Neural $\COs$, where $\CO$ parameters assume the role of latent variables of a hierarchical \textit{variational autoencoder} (VAE) \citep{kingma2013auto}. $\CO$ VAEs are shown to be less sensitive than state--of--the--art VAEs \citep{child2020very} to loss hyperparameters via a rate--distortion analysis \citep{alemi2018fixing}, and can sample at resolutions unseen during training. This is achieved by magnifying images through $\COs$, achieved by decoding at higher resolutions. In dynamically binarized MNIST, data samples of $\CO$ VAE trained on $28\times 28$ images are magnified up to $40 \times$ to a resolution of $1120\times 1120$, revealing additional detail over upsampling via interpolation. Finally, we showcase applications for fractal art, where an image can be "fractalized" i.e. reconstructed as a collage of smaller copies of itself appearing at different scales (see Figure \ref{fig:1mandelbrot}, top).

%% file: sections_arxiv/02_preliminaries.tex
\section{Problem Setting}
Our main goal in this section is to succintly formalize the \textit{fractal data encoding} optimization problem at the heart of fractal compression, as well as our proposed approach. To do so, we lay foundations following \citep{barnsley1985iterated,fisher2012fractal,barnsley2014fractals}.   

\subsection{Background and Notation}
Let $(\bX, d)$ be a complete metric space and $(\cH(\bX), d_{\cH})$ its corresponding Haussdorff metric space, i.e. $\cH(\bX) = \{\bA\subset \bX : \bA~\text{is compact}\}$. We represent a data point as some set $\bS\in\cH(\bX)$. This choice of space supports an application--agnostic treatment of the fractal data encoding problem. A concrete realization will be discussed for image domains. We note that a self--contained reference is provided in Appendix A.

\begin{example}
Consider binary images on a square domain. Then, $\bX$ is a finite compact subset of $\mathbb{R}^2$. An image $\bS$ is the finite set of coordinates of either black or white pixels. Further, each image corresponds to a point in $\cH(\bX)$.
\end{example}
Let $\{f_1, f_2,\dots, f_K\}$ be a collection of maps on $\bX$, $f_k:\bX\rightarrow\bX$. This is colloquially referred to as \textit{iterated function system} (IFS) \citep{barnsley2014fractals}. We can then define a map $F:\cH(\bX)\rightarrow\cH(\bX)$ by
\begin{equation*}
    F(\bA) = \textstyle\bigcup_{k=1}^K f_k(\bA)\quad\forall\bA\in\cH(\bX)
\end{equation*}
where $f_k(\bA)$ is intended as $f_k(\bA) = \{f_k(a) : a\in\bA\}$. 

An interpretation of the above is given by the following: $F$ produces as output a composition, or \textit{collage}, of transformations applied to a subset $\bA$ of $\bX$.  
\subsection{The Inverse Problem: Data to IFS} 

\textit{Given data $\bS$, can we find a map $F$ with $\bS$ as its fixed point?} This can be achieved by identifying a collection of maps $f_k:\bX\rightarrow\bX$ such that the following conditions hold
    \[
        \begin{aligned}
                i.~~& \text{$F:\cH(\bX)\rightarrow\cH(\bX); \bA\mapsto\textstyle\bigcup_{k=1}^K f_k(\bA)$ is contractive};\\
                ii.~~& \text{$\bS$ is \underline{the} fixed point of $F$},~\bS = F(\bS) = \textstyle\bigcup_{k=1}^K f_k(\bS);
        \end{aligned}
    \]
Note that, $F$ is contractive w.r.t the Hausdorff metric with Lipsichitz constant $L<1$ iff all the maps $f_k$ are contractive w.r.t $d$ with constant $\ell_k<1$. In such a case it holds $L = \max_k\{\ell_k\}$. Note that $F$ admits a unique fixed point. 

A classical result provides one constructive path towards an optimization problem to find such $F$.
\begin{theorem}[Collage Theorem (CT) \citep{barnsley1985iterated}]\label{thm:co} Let $(\bX, d)$ be a complete metric space and let $f:\bX\rightarrow \bX$ be a $\ell$--Lipschitz contractive map with fixed point $x^*\in\bX$. Then,
\begin{equation}
    d(x, x^*) \leq \frac{1}{1-\ell}d(x, f(x))
\end{equation}
\end{theorem}

By applying the CT directly to $F$ using the Hausdorff metric $d_\cH$ we have 
\[
    d_\cH(\bS, \bA^*)\leq \frac{1}{1 - L}d_\cH\left(\bS, \textstyle\bigcup_{k=1}^Kf_k(\bS)\right).
\]
This means that we can upper bound the distance $d_\cH(\bS, \bA^*)$ between data $\bS$ and attractor $\bA^*$ of $F$ via $d(\bS, F(\bS))$, which requires a single application of $F$ and is thus cheaper to evaluate. Moreover, even when it is not possible to stitch together transformed copies $f_k(\bS)$ to perfectly reconstruct the data $\bS$, i.e. $d_\cH\left(\bS, \textstyle\bigcup_{k=1}^Kf_k(\bS)\right) \neq 0\quad(\Leftrightarrow \bS\neq F(\bS))$,
a smaller IFS Lipschitz constant $L$ of the IFS implies a lower distance between the data $\bS$ and the attractor $\bA^*$ of $F$, given a mismatch $d_\cH(\bS,F(\bS))$.

This, in turn, implicitly promotes the use of ``very contractive'' maps $f_k$ (i.e. with low $\ell_k$). We refer to the procedure of searching for an $F$ that minimizes the r.h.s of the CT bound as the \textit{fractal data encoding} problem. 

\paragraph{A learning perspective of fractal data encoding} In the language of machine learning, fractal data encoding problem can be translated into finding a parametric representation $f_k(~\cdot~; w_k)$, $w\in\R^{n_w}$ for functions $f_k(\cdot)$ (e.g. neural networks with parameters $w_k$) where $w = (w_1,\dots, w_K)\in\bW$ are optimized to minimize a Hausdorff metric loss function $d_\cH (\bS, F(\bS; w))$ naturally induced by the CT, i.e.
\begin{equation}\label{eq:opt_ct}
        \min_{w\in\bW}~~d_\cH (\bS,\textstyle\bigcup_{k=1}^K f_k(\bS; w_k))
\end{equation}
Once optimal $f_k(\cdot; w_k)$ are obtained, it is possible to find the data that $F$ encodes in parameters $w$ (i.e. the \textit{decoding} process): after sampling any initial condition $\bA_0$, the original data can be decided by iterating $\bA_{t+1} = F(\bA_t)$, until convergence to $\bA^*\approx \bS$. 
\paragraph{Solving for affine IFS}
As with traditional approximation problems, there is a tension in the objective of fractal data encoding between the "expressiveness" of the class of functions, and the tractability of the optimization problem. 

The solution $w$ of \eqref{eq:opt_ct} is an equivalent representation for $\bS$ (up to $d_\cH(\bS, \bA^*)$). However, the choice of parametrization of $F$ should be informed by  a downstream task where $w$ ought to be used. Indeed, a general fractal data encoding problem \eqref{eq:opt_ct} is \textit{task--agnostic}. 

Existing methods based on the idea of \citep{barnsley1985iterated} resolve this tension by considering (a) compression as a task, such that $w$ should be encodeable in the least number of bits possible and (b) affine functions $f_k(x; w_k) = a_k x + b_k$. With these choices, a solution to \eqref{eq:opt_ct} can be found in closed--form \citep{fisher2012fractal}, and the parametrization $w$ results compact enough to be a valid compression code -- only two floats for each $f_k$ in $F$, i.e. $w_k = (a_k, b_k)\in\R^2$.

There are a number of limitations we aim to address:
\begin{itemize}
    \item Fractal data encoding is only considered as an intermediate step towards compression. However -- as certified by machine learning practice -- a data representation is only as useful as the tasks it allows to solve. We develop a learning--based approach to the solution of \eqref{eq:opt_ct} for tasks beyond compression.
    \item Solving \eqref{eq:opt_ct} on a collection of data as per \citep{fisher2012fractal} is computationally expensive, even when a closed--form solution is ensured by a restriction to affine IFSs. We directly solve a collection of fractal data encoding problems in parallel via hypernetworks \citep{ha2016hypernetworks}, effectively amortizing the cost.
    \item As noted by \citep{welstead1999fractal,fisher2012fractal}, for an (affine) IFS to provide a satisfactory solution to \eqref{eq:opt_ct}, the self-similarity property has to be global across the set $\bS$. That is, the entire set $\bS$ is made up of smaller copies of itself, or a part of itself, property that is rather rare in natural data: indeed, most images are only self--similar to a degree. To alleviate these restrictions, we develop $\CO$ operators, a generalization of IFSs which can be broadly categorized as a soft--\textit{partitioned iterated function system} (PIFS) \citep{jacquin1992image}.
\end{itemize}

\subsection{IFS, PIFS and Beyond}
The limited approximation capabilities of IFSs lead to the development of more general classes, most notably \textit{partitioned iterated function systems} (PIFS) \citep{jacquin1992image}. PIFS can capture localized self--similarity by allowing each domain of a contraction map $f_k$ to be a different subset $\bA_k \subset \bS$. This introduces a significant challenge in \ref{eq:opt_ct}: the optimization problem need now determine optimal (as measured by $d_\cH$) domains $\bA_k$ for each $f_k$ by searching across all possible subsets of $\bS$, yielding an exploding combinatorial problem. In practice, this entails a choice of type of subsets (or \textit{partition}) to search over\footnote{Note further that a compression code of PIFS requires storing an address of the domain of each $f_k$, other than the parameters $w_k$. This limits the type of subset allowed to those that support "short" parametrizations.}.    

By construction, Neural $\COs$ will be shown to provide a direct solution to the combinatorial problem of optimal domain search for PIFS. This is to ensure a Neural $\CO$ can be seamlessly trained end--to--end. 

%% file: sections_arxiv/03_collage_ops.tex
\section{Neural $\COs$}
\label{sec:CO}
Moving forward, we treat Neural $\COs$ algebraically. This allows us to discuss in detail the properties of a Neural $\CO$, including differences with a PIFS. To do so we consider, instead of generic sets, data that can be expressed as simple ordered sets: in other words, as vectors. Images will be our recurring example, with the understanding that the entire discussion can readily be adapted to other modalities e.g. sequences. 
\begin{wrapfigure}[18]{r}{0.5\textwidth}
    \vspace{5mm}
    \centering
    \includegraphics[scale=.95]{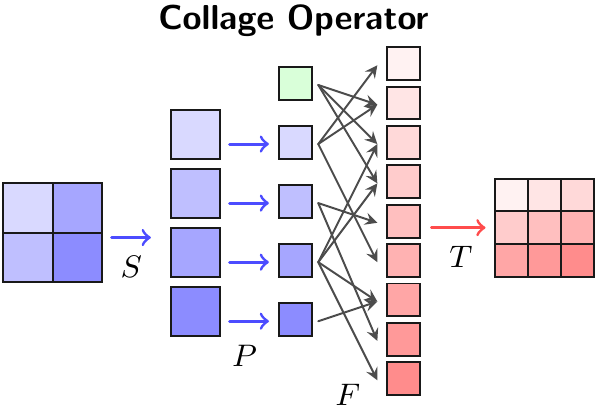}
    \vspace{-1mm}
    \caption{Conceptual schematic of a $\CO$. In blue, domain cells $\sS_n$; in red, range cells $\sR_k$ of ranges. Green highlights auxiliary domains. A step can be broken down into (1) $S$ partitions into domains (2) $P$ reduces dimensions to ensure dimensions match (3) $F$ produces all range cells, each following \eqref{eq:step} (4) $T$ rearranges the output.}
    \label{fig:ex_co}
    \vspace{-4mm}
\end{wrapfigure}

\paragraph{From generic sets to vectors}
Following the PIFS treatment of \cite{10.5555/191581.191600}, we focus our analysis on $\CO$ composed of \textit{affine} maps, operating on the space of discrete images of a given resolution with a total number $m$ of pixels each taking values in $\R$. Pixels of different channels are treated without loss of generality as different elements. This allows us to collect all $m$ pixel values in an \textit{ordered}\footnote{with a specific predefined criterion, e.g. row--major ordering.} vector $z\in\R^m$. 

A type of subsets on images involves the formation of square patches. Let us assume that each image is partitioned into (1) $K$ non--overlapping \textit{range cells} and (2) $N$ possibly--overlapping \textit{domain cells}. A range cell $\sR_k$ is then of size $n_{r,k}\times n_{r,k}$, such that $m = \sum_{k=1}^K n_{r,k}^2$, and domain cells $\sD_n$ are of size  $n_{d,n}\times n_{d,n}$.

With such coordinatization, the (affine) fixed--point map $F$ reduces to a linear operator on $\R^m$. This \textit{discrete} representation allows deriving $\CO$ operators in an algebraic form, amenable to a practical realization in a learning algorithm.
\begin{defn}[Neural $\CO$ Operator]\label{def:nco}
    Consider a $m$-pixel image represented by the ordered vector $z\in\R^m$. Then, a \textit{Collage Operator} is defined as the parametric linear map:
    \begin{equation} \label{eq:co_vanilla}
     F(z; w) = \sum_{k, n}{\color{orange!80} \gamma_{k,n}}{\color{blue!70}a_{k,n}} T_k P_{k,n} S_n z + \sum_{k,n}{\color{orange!80} \gamma_{k,n}} {\color{blue!70}b_{k,n}} T_k \1
     \vspace{-4mm}
    \end{equation}
    %
    %
    \begin{itemize}
        \item $S_n\in\R^{n_{d,n}^2\times m}$ \textbf{selects} a \textit{domain cell} $\sD_n$ of $n_{d,n}\times n_{d,n}$ pixels.
        \item $P_{k,n}\in\R^{n_{r,k}^2\times n_{d,n}^2}$ is a \textbf{pooling} operator that shrinks the domain cell $\sD_k$ into the size of the corresponding range cell $\sR_k$, i.e. from $n_{d,n}\times n_{d,n}$ to $n_{r,k}\times n_{r,k}$ pixels;
        \item $T_k\in\R^{m\times n_{r,k}^2}$ \textbf{positions} the pooled domain cell in the correct range cell location and zeroes out the rest;
        \item ${\color{blue!70}a_{k,n},b_{k,n}}\in\R$ \textbf{scales} and \textbf{translates} the value in each pixel of the pooled domain cell, respectively.
        \item ${\color{orange!80}\gamma_{k,n}}\in\R$; convex \textbf{combination} of affine outputs produced from all $N$ domains. 
    \end{itemize}
    The parameters $w$ is the collection of all $a_{k,n}, b_{k,n}$ and the mixing weights $\gamma_{k,n}$.
\end{defn}
$\CO$ operators represent each range cell $\sR_k$ as a convex combination of pooled and scaled versions of all domain cells $\sD_n$ translated block--wise by $b_n$. On individual range cells comprising the output, a symbolic representation can be given as
\begin{equation}\label{eq:step}
    \sR_k = \sum_{n} \gamma_{k,n} a_{k,n} \sD_n + \sum_{n} \gamma_{k,n} b_{k,n}. 
    \vspace{-1mm}
\end{equation}
In the generic set formulation, these maps correspond to functions $f_k$ of $F$. This highlights the first major difference with a PIFS: each $f_k$ \underline{does not} act on a different subset (domain cell) to produce $\sR_k$. Instead, all $f_k$ aggregate affine transformations -- parametrized by $a_{k,n}, b_{k,n}$ -- on domains via $\gamma_{k,n}$. However, each $f_k$ is equipped with different mixing weights and different affine maps. Hence, a $\CO$ in this form can be seen as a soft--PIFS. %

\paragraph{Introducing auxiliary domains}
A $\CO$ step maps mixtures of all domains to each range, and assembles the ranges into its output. As Neural $\COs$ are often optimized on datasets, rather than single data points, we posit that improvements in the expressiveness can be readily achieved by mixing additional dataset--level information through \textit{auxiliary domains} $\sU_v$.

\begin{defn}[$\CO$ operator with auxiliary domains]
    \[
        \hat{\sR}_k = \sR_k + \sum_{v=1}^{V}\gamma_{k,v}a_{k,v}\sU_v 
    \]
    In coordinates this translates to 
    \begin{equation*}
        \begin{aligned}
            \hat{F}(z, u; w) &= F_w(z; w) + \sum_{k=1}^K \sum_{v=1}^V{\color{orange!80} \gamma_{k,v}}{\color{blue!70}a_{k,v}} T_k P_{k,v} S_v u  \\
        \end{aligned}
    \end{equation*} 
    where $F_w(z)$ is \eqref{eq:co_vanilla} and $\sR_k$ is \eqref{eq:step}, and the parameters $w$ include the coefficients $\gamma_{k,v},a_{k,v}$ of the auxiliary domains $\sU_v$, as well as $\gamma_{k,n},a_{k,n},b_{k,n}$. 
\end{defn}

We consider different variants of $\sU_v$, including: deterministic transformations of domain cells $\sD_n$ e.g. rotations as per \citep{jacquin1992image}, learned cells directly parametrized and optimized for an objective, similar to feature maps in \citep{jaegle2021perceiver}, and $\sU_v$ produced by a neural network encoder. Specifics are provided in Section $4$.

A schematic of a single step of $\CO$ is given in Fig.~\ref{fig:schematic}.

\subsection{The Forward Problem: Collage to Data}
Given a parametrization $w$ for the $\CO$ operator $F_w$ and an initial image $z_0$, the attractor $z^*~:~z^* = F_w(z^*)$ can be recovered by iterating the fixed--point map
\[
    {z_{t+1}} =F(z; w) = A(w)z_t + b(w)
\]
assuming $F$ to be a contraction w.r.t. the standard Euclidean metric on $\R^m$. This can be ensured by an appropriate choice of the coefficients $a_{k,n}, \gamma_{k,n}$. In particular, if all the mixing weights are such that $\sum_{n=1}^N \gamma_{k,n} = 1$ and $|a_{k,n}| < 1$, then contractivity of the collage operator follows as in standard PIFS (see e.g. \cite{fisher2012fractal}).

\begin{wrapfigure}[13]{r}{0.5\textwidth}
    \vspace{-5mm}
    \centering
    \includegraphics[width=\linewidth]{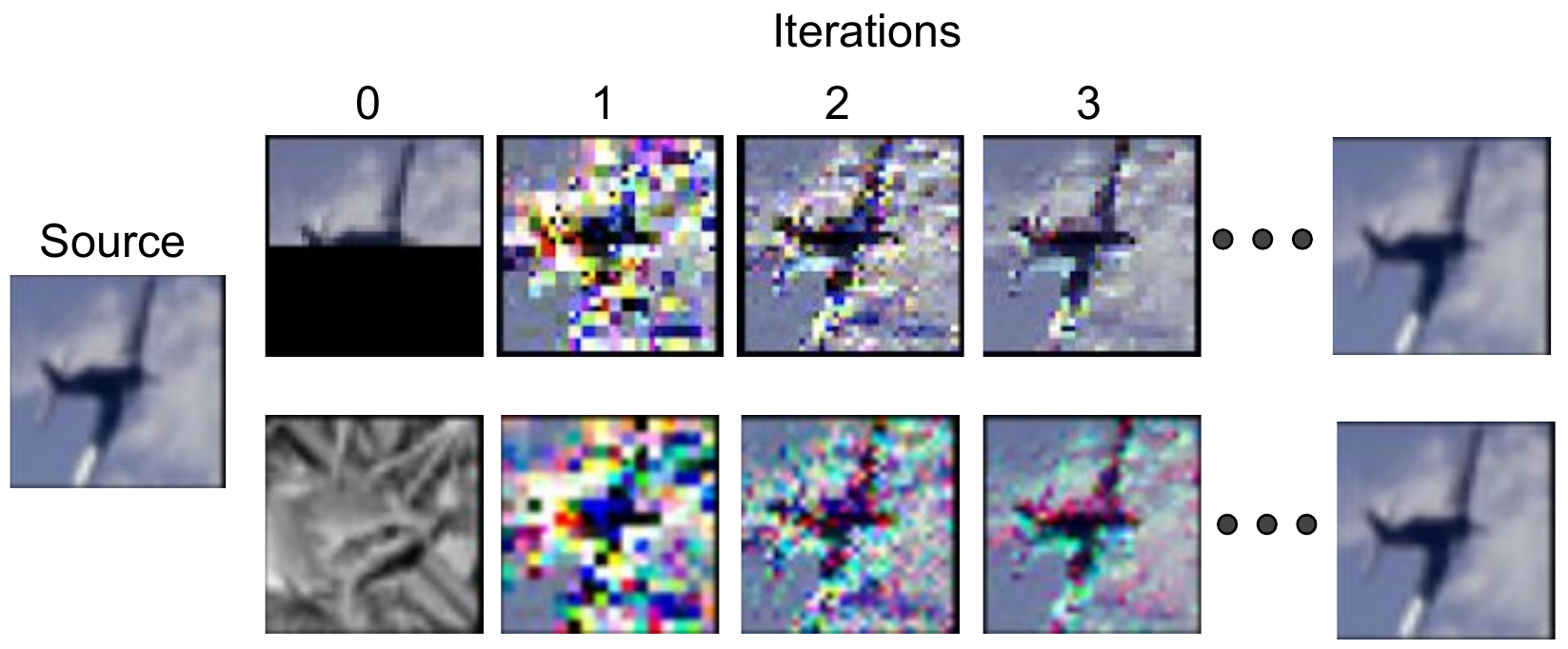}
    \vspace{-3mm}
    \caption{Forward problem of a $\CO$: given $w$, different initial conditions lead to the same decoded data $z^*$.}
    \label{fig:example_co}
    \vspace{-6mm}
\end{wrapfigure}

Note that the attractor of a $\CO$ can be also computed in closed--form as $z^* = [\Id - A(w)]^{-1}b(w)$. A similar discussion follows for a general $\CO$ with auxiliary domains. Note that auxiliary domains $\sU$ across iterations $t$ are to be chosen such
that the sequence $\{\sU_t\}_{t=0}^{\infty}$ converges e.g. constant functions. Figure \ref{fig:example_co} provides a visualization of the convergence of a $\CO$ to its fixed--point.
\paragraph{Decoding at higher resolutions}
A $\CO$ can be applied, without change, to images of different resolutions. Consider scaling the resolution by a factor $s>1,~(s\in\bN)$. Then the \textit{magnified} image representation is made up of $s^2$ images $z^i\in\R^m$. The $\CO$ operator can then be thought to act singularly on each $z^i$ obtaining a forward fixed--point iteration 
\[
    z^i_{t+1} = \sum_{k, n}{\gamma_{k,n}^i}{a_{k,n}^i} T_k P_{k,n} S_n z_t^i + \sum_{k,n}{\gamma_{k,n}^i} {b_{k,n}^i} T_k \1.
\]
When solving by unrolling the fixed--point iteration, the second term can be precomputed as it does not depend on $z_t$.

\subsection{Amortized Solution of the Inverse Problem}

Although the CT suggests a constructive procedure via \eqref{eq:opt_ct} to find a valid fractal representation $w$, there are no guidelines in case other objectives are of interest. Further, the class of PIFS -- without modifications -- does not lend itself well to numerical optimization as it involves the combinatorial problem of matching domains to ranges. With their soft aggregation, Neural $\COs$ can instead be used for task--based optimization.

Given an input image $x\in\R^m$, we can optimize the parameters $w$ (and pixel values $u$ of the auxiliary domains) of the Neural $\CO$ operator to minimize an objective $J_\omega(x, z^*(w, u))$ by solving a nonlinear program $\min_{w, u} J_w(x, z^*(w, u))$ with $z^*(w, u)$ obtained by the fixed point iteration on $\hat F(z, u; w)$. Choosing $J_w$ to be a reconstruction objective yields a problem similar to fractal data encoding as defined by \eqref{eq:opt_ct}.

Suppose instead to be given an image dataset whose distribution $p$ is known only through i.i.d. samples $x\in \R^m,~x\sim p(x)$. In this context, the fractal data encoding problem in standard form needs to be solved for each sample $x$. Instead, we introduce an \textit{hypernetwork} \citep{ha2016hypernetworks} $\cE$ with weights $\theta$, generating a set of coefficients $w$ given an input image $x$, i.e. parametrizing a map $x \mapsto w_\theta(x)$,
\[
    \forall x\sim p(x)~~~w_\theta(x) = \cE(x; \theta)
\]

The hypernetwork is trained to solve the following empirical risk minimization problem, effectively amortizing the cost over the full dataset:
\begin{equation}\label{eq:inv_prob}
    \begin{aligned}
        \min_{\theta, u}~~&\bE_{x\sim p(x)}\left[J(x, z^*(\theta, u))\right]\\
        \text{subject to}~~ & w_\theta(x) = \cE(x; \theta)\\ 
        &z^*(\theta, u) = \hat F(z^*(\theta, u), u; w_\theta(x))\\
        & u\in \bU.
    \end{aligned}
    \vspace{-1mm}
\end{equation}
To optimize Neural $\COs$ in general non--encoding tasks, the objective can be adapted in example as $J(y, z^*(\theta, u))$, where $y$ is the label corresponding to $x$.

%% file: sections_arxiv/05_experiments.tex
\section{$\COs$ in Learning Tasks}
\begin{wrapfigure}[15]{r}{0.48\textwidth}
    \vspace{-2mm}
    \centering
    \includegraphics[scale=1]{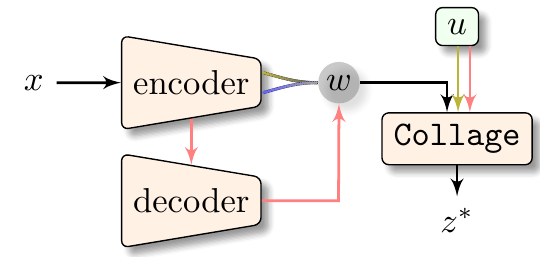}
    \caption{Schematic of a Neural $\CO$. In {\color{yellow!60!black}yellow}, computation unique to the compressor variant; $x$ is deterministically encoded into $\CO$ parameters $w$, then used to decode. Similarly, {\color{blue!40!black}blue} indicates computation done by the fractalizer, and {\color{red!40!black}red} by the generative variant, where $w$ takes the role of a latent variable.}
    \label{fig:schematic}
    \vspace{-4mm}
\end{wrapfigure}

We showcase applications of Neural $\COs$ for fractal art, image compression and generation. All variants of the model share a common structure, outlined in Figure \ref{fig:schematic}. 


%
\subsection{Neural $\COs$ for Fractal Art}

When data is represented through the parameters $w$ of a $\CO$, it can be arbitrarily magnified by decoding at any resolution. The type of patterns revealed through magnification need not be corresponding to real detail missing from the image. In particular, the patterns found depend on how one generates domains, auxiliary domains and class of operator $F_w$. A similar phenomenon has been observed in the literature of fractal compression \citep{mitra2000technique}.

\begin{figure}[t]
    \centering
    \includegraphics[width=0.6\columnwidth]{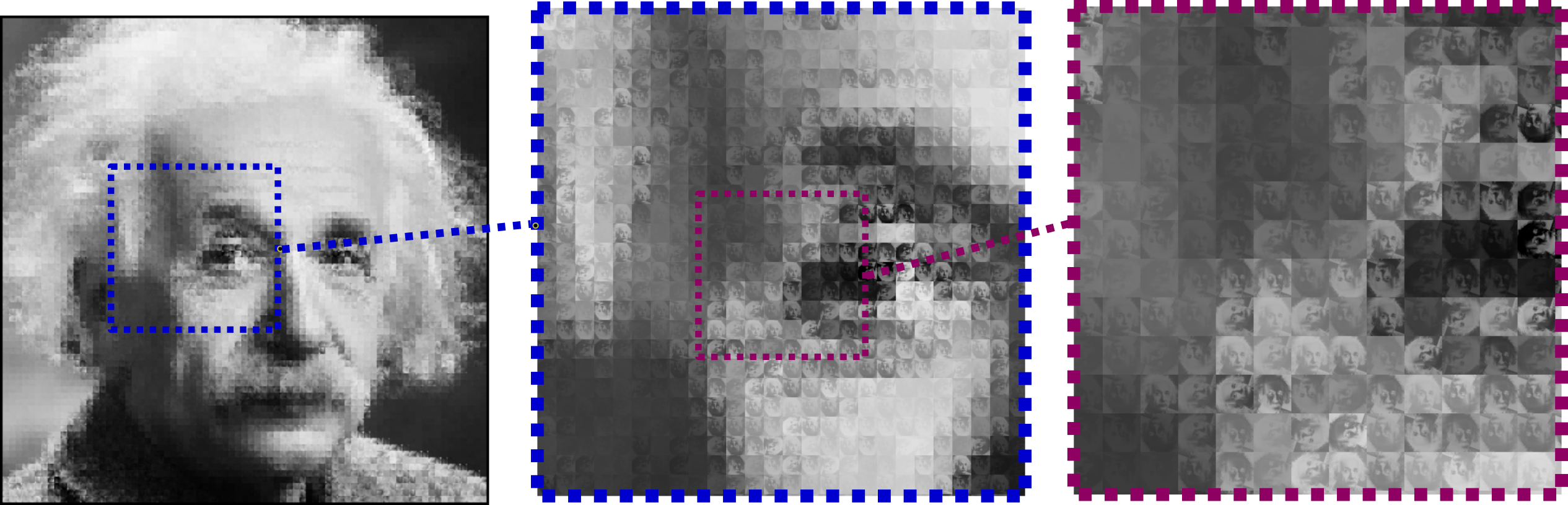}
    \includegraphics[width=0.6\columnwidth]{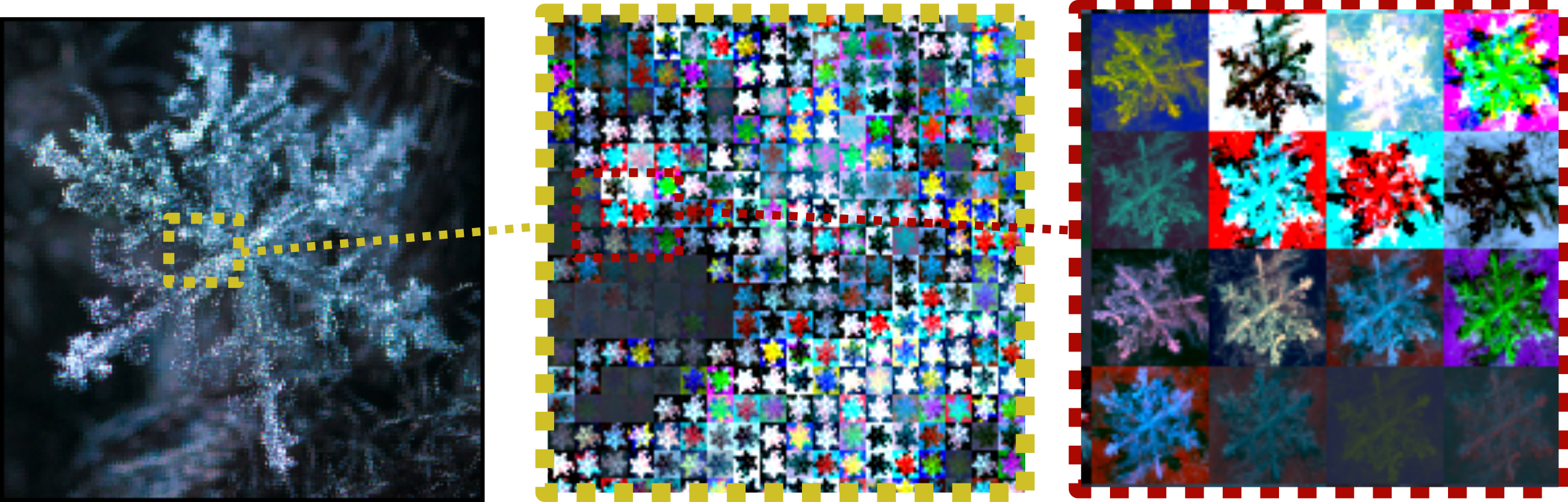}
    \vspace{-1.5mm}
    \caption{Fractalization of greyscale and RGB images via $\CO$. The images are compressed as the coefficients of a $\CO$, and decoded as its attractor. Note that the above is a showcase of \textit{global} fractalization: no partitioning is performed, and the entire image itself is taken as the source.}
    \vspace{-2mm}
    \label{fig:art}
\end{figure}

As an example, consider Figure \ref{fig:art}, where the fractal pattern of snowflakes within snowflakes does not correspond to reality. We call this type of globally self--referential magnification as fractalization of an image. Here, we use Neural $\CO$ fractalizers as a means to generate visual fractal art. To this end, we seek fractalization such that magnification in any region of the image yields the same target (up to affine transformations of it). This allows the creation of animated loops where the fractalized image is gradually magnified by decoding at increasing resolutions.

\paragraph{Experimental details}
We solve the inverse problem of a $\CO$, namely image $x$ to $w$, via a convolutional architecture parametrized by $\theta$, loosely based on {\tt ConvMixer} \citep{trockman2022patches}. The objective of the inverse problem \eqref{eq:inv_prob} is a reconstruction objective $J(x, z^*(\theta, u)) = \sum_{i=1}^m (x_i - z_i^*(\theta, u))^2$ where $z^*$ is the fixed--point of the Neural $\CO$ with parameters $w$ computed through the encoder $w = \cE_\theta(x)$. We choose the single domain $\sD$ to be the entire image. Before applying $F_w$, we augment the domain via rotations of itself, utilizing those as auxiliary domains.   

%
%

%
Figure \ref{fig:1mandelbrot} and \ref{fig:art} show example fractalizations possible with Neural $\COs$, on greyscale and RGB images. The images can be magnified to any resolution (up to memory limits), revealing multiple fractal levels. Additional details and implications are reported in the Appendix. 

\subsection{Neural $\CO$ Compressors}

Next, we apply Neural $\CO$ to image compression. We store images as the parameters $w$ of an affine $\CO$ produced by an encoder, this time trained on a dataset. After training, the encoder can be used to compress additional images in parallel with a single forward pass by producing the corresponding $w$, thus amortizing the cost of solving the inverse problem. Figure \ref{fig:schematic} provides an overview of the computation done by a Neural $\CO$ compressor. 

Differently from Neural $\COs$ fractalizers, compressors employ learned feature maps as auxiliary domains. Such domains are directly parametrized and optimized in pixel--space and match range cells in dimension as to avoid unnecessary pooling.  

\begin{wraptable}{r}{0.58\textwidth}
    \centering
    \begin{tabular}{c|cc}\toprule
        \textbf{Method} & \multicolumn{2}{c}{\textbf{$\uparrow~$PSNR$~|$~bpp~$\downarrow$}}\\
        \midrule
        \rowcolor{red!4}
        Fractal (no aug.) & $31.06~|~0.14$&  $31.68~|~0.36$ \\
        \rowcolor{red!6}
        Fractal (augment.) & $30.51~|~0.15$&  $30.80~|~0.39$ \\
        \rowcolor{blue!6}
        COIN & $25.74~|~0.17$& $27.34~~|~0.33$\\
        \rowcolor{blue!4}
        Neural $\CO$ \textbf{(ours)} & $31.30~|~0.13$& $32.12~|~0.31$ \\
        \rowcolor{green!4}
        block--DCT & $33.22~|~0.13$& $34.65~|~0.32$
    \end{tabular}
    \caption{Average \textit{peak signal-to-noise ratio} (PSNR) at low ($\approx 0.15$) and medium ($\approx 0.30$) \textit{bits--per--pixel} (bpp) budgets of baselines ({\color{red!50}baseline fractal}, {\color{blue!50}implicit}, {\color{green!30!black}spectral}) and Neural $\COs$ compressors. Results on held--out on $1200\times 1200$ crops of the DOTA dataset \citep{Xia_2018_CVPR}. PSNR of Neural $\COs$ introduces less visible artifacts than other self--similarity or implicit compression schemes, narrowing the gap with spectral compression methods such as JPEG.}
    \label{tab:comptab}
    \vspace{-9mm}
\end{wraptable}
For compression, the main desideratum is visual fidelity: regular domains $\sD_n$ allow Neural $\COs$ to capture intra--image self--similarity, whereas auxiliary domains $\sU$ optimized for image quality complement them by focusing on inter--image patterns.

\begin{figure*}[t]
    \centering
    \includegraphics[width=\textwidth]{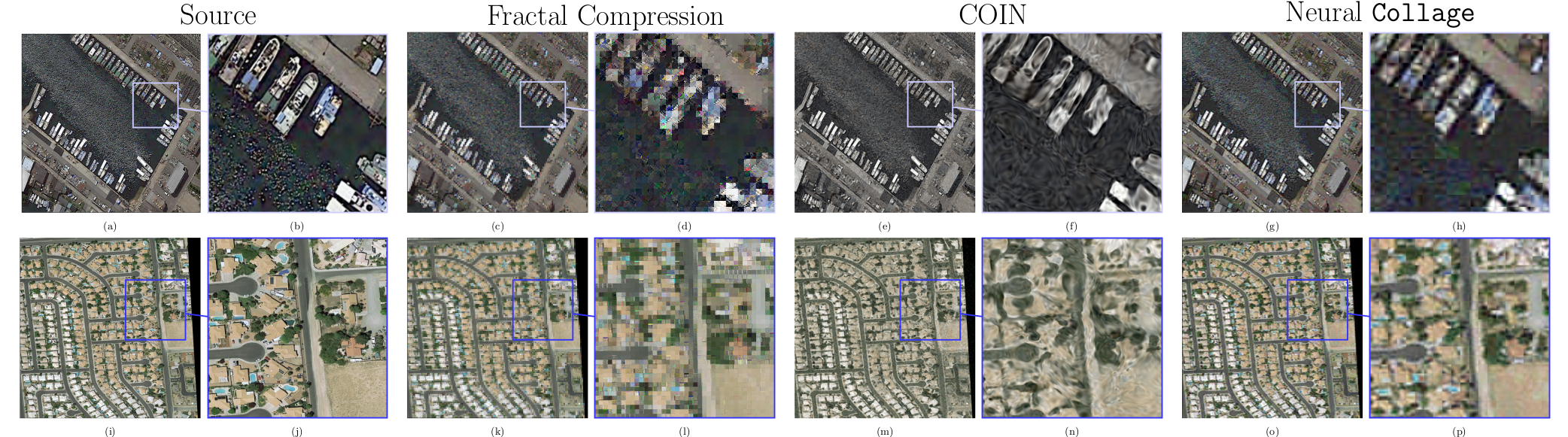}
    \vspace{-3mm}
    \caption{Visual comparison of decoded images from $1200\times 1200$ crops of the DOTA dataset, obtained from different compression methods ($\approx 0.30$) bpp. Images decoded from Neural $\CO$ codes exhibit less noticeable artifacts, with improved color preservation over COIN and more detail than fractal compression.}
    \label{fig:compfig}
\end{figure*}
\paragraph{Compression of high--resolution aerial images}
We consider compressing images obtained from the DOTA  large-scale aerial images dataset \citep{xia2018dota}. From the DOTA training set, we produce $80000$ random $40\times 40$ crops as our training dataset. We use the same encoder architecture as for Neural $\CO$ fractalizers. We optimize encoder parameters $\theta$ and auxiliary sources on the reconstruction objective $J(x, z^*(\theta, u)) = \sum_{i=1}^m (x_i - z_i^*(\theta, u))^2 + \|w\|_2^2
$ The model is trained on the $40\times 40$ crops and evaluated on $10$ held-out $1200\times 1200$ images. This is possible as an image of any resolution can be first broken up into blocks of appropriate size, in this case the training resolution, $40 \times 40$, passed through the encoder to obtain the corresponding parameters, then concatenated to construct a valid code for the entire image. This operation can be performed in parallel by treating each block as an element of a pseudo--batch. Thus, Neural $\CO$ compressor can thus be used to compress images of different resolutions at test time.
\begin{wrapfigure}[18]{r}{0.52\textwidth}
    \centering
    \includegraphics[scale=0.95]{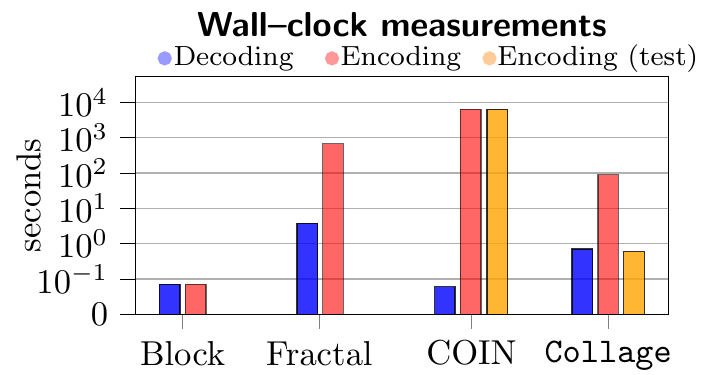}
    \vspace{-2mm}
    \caption{Wall--clock time (s) measurements of compression methods applied to the held--out set of $10$ DOTA ($1200\times 1200)$. For neural compressors, we report encoding times including training, as well as encoding times on unseen images (after training). At test time, encoding for Neural $\CO$ compressors is $100\times$ faster than fractal compression.}
    \label{fig:wallclock}
    \vspace{-6mm}
\end{wrapfigure}

\paragraph{Results}
We compare \textit{peak signal-to-noise ratio} (PSNR), in addition encoding and decoding measurements for a variety of compression baselines. We contextualize our results with comparisons to both non--neural as well as another implicit neural compressor. In particular, we evaluate the performance of a standard PIFS--based fractal compression as per \citep{jacquin1993fractal}, implemented to exploit parallelization on GPU, and COIN \citep{dupont2021coin}. Fractal compression baselines and $\CO$ both use non--adaptive tiling partitioning schemes. We evaluate two variants of fractal compression, one where domain cells are augmented via rotations and color flips \citep{welstead1999fractal}, and one without augmentations. We further compare with block--DCT, the spectral lossy compression backbone of most JPEG codecs. All compression methods are evaluated at low and medium bpps, with metrics provided in Table \ref{tab:comptab}. Figure \ref{fig:compfig} provides a visual comparison of images obtained by decoding the lossy code of each method. Neural $\COs$ show less noticeable artifacts and improved color retention. 

Finally, Table \ref{tab:comptab} provides wall--clock time measurements of all methods during the respective per image encoding and decoding procedures. Results for COIN and Neural $\CO$ measure encoding times (including the training procedure), and encoding times after training. Neural $\COs$ are orders of magnitude faster than fractal compression and, at test--time, of COIN. Although spectral lossy compressors common in state--of--the--art codecs perform with best PSNR in medium and high bpp settings, $\COs$ narrows the gap in terms of reconstruction quality as well as encoding speed.

\paragraph{Code length of a $\CO$}
The total \textit{bits--per--pixel} (bpp) cost of the code $w$ corresponding to the parameters of a $\CO$ depends on the coding scheme used to store each numeric entry. We exploit bounds on $a, b, \gamma$, enforced via ${\tt tanh}$ and softmax, as well as regularization to reduce the total cost. The $L_2$ regularization term on $w$ is introduced to shrink the range of values assumed by elements of $w$, ensuring that less bits can be used for storage. We do not use lossless coding schemes to store parameters of $\COs$ and other baselines. Further details on bbp computation are provided in the Appendix.

%




\subsection{Generative Neural $\COs$}
Next, we investigate application of Neural $\COs$ as deep generative models for distributions $p(x)$ of images. In this context, $w$ assume the role of latent variables. We consider 
a \textit{variational autoencoder} (VAE) \citep{kingma2013auto} model based on Neural $\COs$, noting that other classes e.g. diffusion models \citep{song2020score,kingma2021variational}  may be used instead. More specifically, we choose a hierarchical VAE as our starting point, VDVAEs \citep{child2020very}, a state--of--the--art architecture for VAEs. For compactness, the objective will be described using a single--level $\CO$ VAE. For a modern extension to the hierarchical case, we refer to \citep{vahdat2020nvae,child2020very}.

\paragraph{Magnifying samples via $\CO$ VAEs}

VAE models seek to data $x$ into a latent representation $w$ such that the following lower bound ({\tt ELBO}) on data log--likelihood be maximized
\[
    \bE_{q_\phi}[\underbrace{\log p_\theta (x|w)}_{-\text{distortion}}] - \underbrace{D_{KL}[q_\phi(w|x)||p_\phi(w)]}_{\textit{rate}} \leq \log p(x;\theta,\phi)
\]
where approximate posterior $q_\phi(w|x)$, prior $p_\phi(w)$, and generator $p_\theta(x|w)$ are implemented as neural networks. A multiplicative hyperparameter $\beta$ is often introduce to control the relative weight between rate and distortion \citep{higgins2016beta}. In a $\CO$ VAE, $q_\phi(w|x)$ and $p_\phi(w)$ are parameterized as per \citep{child2020very}, except the approximate posterior $q_\phi(w|x)$ need not produce large feature maps but scalar $\CO$ parameters $w$. Moreover,  $p_\theta(x|w)$ is a $\CO$ such that by leveraging the consideration of Section 3.1, samples can be decoded at any resolution.

\begin{figure}[t]
    \centering
    \includegraphics[width=0.57\linewidth]{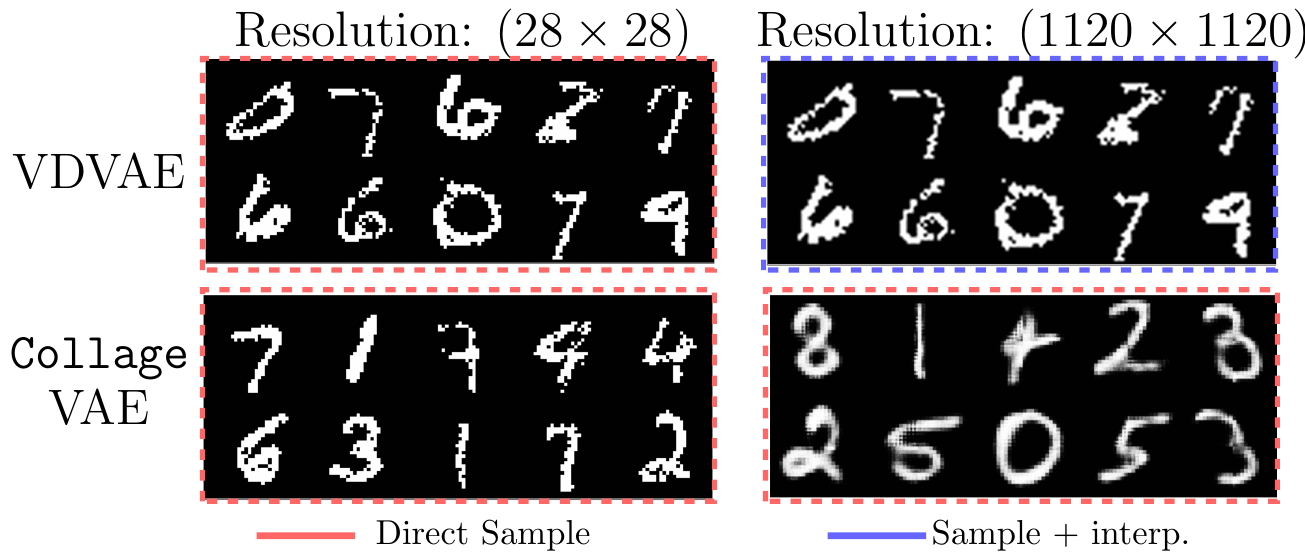}
    \hfill
     \includegraphics[width=0.4\linewidth]{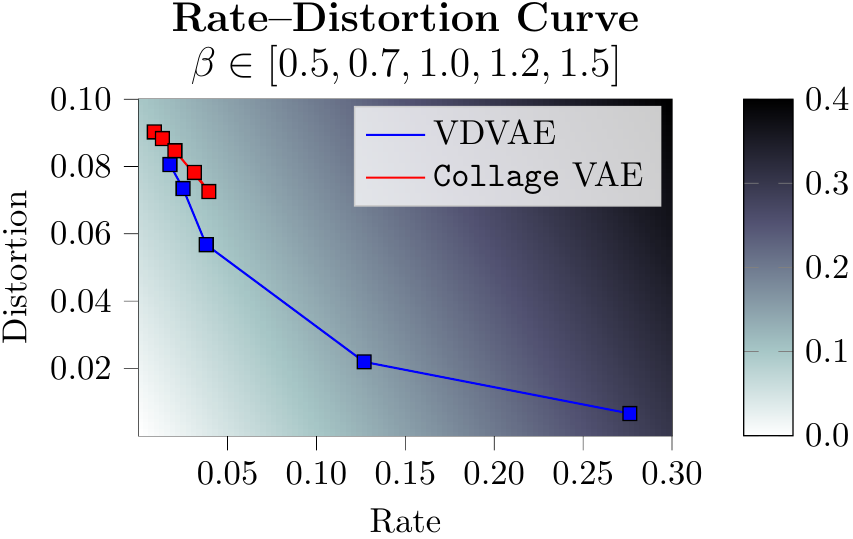}   
    \vspace{-1mm}
    \caption{\textbf{[Left]} Samples obtained from VDVAEs and $\CO$ VAEs on dynamically binarized MNIST. $\CO$ VAEs can generate samples at resolutions unseen during training, adding detail that cannot be obtained by (1) sampling first (2) upscaling via bicubic interpolation, as done in the first row. \textbf{[Right]} Rate--distortion curve on dynamically binarized MNIST, with a curve obtained by sweeping $\beta$ in a range. $\CO$ VAEs are less sensitive to the tuning of $\beta$, and pay a marginal price in Pareto suboptimality to gain the ability to sample at different resolutions.}
    \label{fig:samples}
    \vspace{-4mm}
\end{figure}

\paragraph{Results}
To investigate the properties of $\CO$ VAEs, including quality of magnified samples, we train and compare VDVAEs and $\CO$ VAEs on dynamically binarized MNIST. Rather than aggregate log--likelihoods, we report both test rate and distortion, following a full rate--distortion analysis as described in \citep{alemi2018fixing}. This supports a more detailed evaluation of model behavior at different $\beta$ weights for the rate. We train several $\CO$ VAEs and VDVAEs with $\beta$ ranging in $[0.5,0.7,1.0,1.2,1.5]$, and report the rate--distortion curve in Figure \ref{fig:samples} (right). $\CO$ VAEs are only marginally Pareto suboptimal, but are shown to be less sensitive to training under different $\beta$, which is a common strategy employed to stabilize VAE training ("KL warmup"). Furthermore, $\CO$ VAEs can generate samples at resolution unseen during training, as showcased in Figure \ref{fig:samples} (left). Direct samples at a magnification factor of $40\times$ reveal additional details over VDVAE samples that are magnified via bicubic interpolation. By an appropriate combination of $\CO$ and deep generative models, one may be able to similarly perform zero--shot magnification in other domains. We note that as discussed in Section 5.1, the detail introduced by magnification is entirely dependent on the $\CO$ class; future designs may be developed to introduce types of detail expected in a given dataset.

%% file: sections_arxiv/06_related_conclusions.tex
\section{Related Work and Discussion}
\paragraph{Implicit representations and models}
Representation of data implicitly through functions is extensively used in simulation \citep{osher2004level}. \citep{sitzmann2020implicit,mildenhall2020nerf,dupont2021coin} parametrize the implicit functions via neural networks for use in downstream tasks such as compression. Implicit models, on the other hand, solve optimization problems within their forward pass \citep{poli2020torchdyn}. Neural $\COs$ belong to both classes of methods, being a fixed--point iteration whose parameters define data implicitly. In particular, Neural $\COs$ can be framed as a compactly--parametrized operator variant of \textit{deep equilibrium networks} (DEQs) \citep{bai2019deep}, with parameters produced by a hypernetwork \citep{ha2016hypernetworks}. \cite{huang2021textrm} uses implicit models as implicit representations, with computational advantages gained via fixed--point tracking \citep{massaroli2021differentiable}. Further speedups for Neural $\COs$ compressors could similarly be found via tracking during training. We note concurrent work on an improved version of COIN \citep{dupont2022coin++} that is trained in parallel on patches, similarly to Neural $\CO$ compressors.
\paragraph{Attention operators and patches}
There exist superficial similarities between $\COs$ and attention operators \citep{vaswani2017attention} of vision and language models. In particular, recent variants of vision transformers \citep{dosovitskiy2020image} where attention acts on square patches, can be seen as a single step of a $\CO$, where source and target partition match and aggregation weights are found via similarity scores. $\COs$ differ from attention in that they are structured fixed--point iterations, are built to accommodate non--overlapping partitions, are resolution--invariant, and have a compact parametrization that can be used as a compression code. It remains to be seen whether investigating attention operators through the lenses of $\COs$ can yield improvements in theoretical understanding or performance.

\paragraph{Fractal compression}
The idea of representing images through \textit{iterated function systems} (IFS) dates back to \citep{barnsley1985iterated,barnsley1986fractal}. \cite{jacquin1992image} introduces more flexible fractal compression schemes for images based on \textit{partitioned iterated function systems} (PIFSs). Since then, alternative partitioning schemes i.e. adaptive quadtrees have been proposed. We refer to \citep{fisher2012fractal} for an overview of the main variants. Adaptive quadrees have also been successfully introduced in Transformer architectures \citep{tang2022quadtree}, suggesting that further techniques linked to $\COs$ and fractal compression may be beneficial in this domain. Finally, images generated from IFSs have been used to construct artificial pretraining datasets for large vision models \citep{kataoka2020pre}. Additional references are provided in the Appendix.

\vspace{-1mm}
\section{Conclusion}
This work builds a framework for a learning--based approach to automated discovery of self--similarity in data. We introduce Neural $\COs$, operators equipped with a self--similarity prior, designed to represent data through the parameters of a structured fixed--point iteration. We envisage future application of Neural $\CO$ to other data modalities with naturally occurring self--similarity, such as audio, sequences, or turbulent flows. Beyond additional modalities, neural network modifications informed by the theory of (partitioned) iterated functions systems and Neural $\CO$ may have a further role to play in mainstream learning tasks.
\begin{center}
    \small \textit{Given time, technique never fails. \citep{hewg2022smithing}}
\end{center}

%% file: appendix_arxiv/B_problem_setting.tex
\section{Background and Extended Formulation}
\subsection{Metric Spaces}
\begin{lemma}[Useful results on bounded and closed sets]
    The following hold:
    \begin{itemize}
        \item[$i.$] Let $\bX = \bigcup_{i=1}^n \bA_i$ such that $\bA_i$ is a \textbf{closed} subset of $\R^n$ for all $i = 1,\dots,n$. Then, $\bX$ is \textbf{closed}.
        \item[$ii.$] Let $\bX = \bigcup_{i=1}^n \bA_i$ such that  $\bA_i$ is a \textbf{bounded} subset of $\R^n$ for all $i = 1,\dots,n$. Then, $\bX$ is \textbf{bounded}.
        \item[$iii.$] Let $f:\R^n\rightarrow\R^n$ to be a continuous function and let $\bA$ to be a closed, bounded subset of $\R^n$. Then $f(\bA) = \{f(x): x\in\bA\}$ is also closed and bounded. 
    \end{itemize}
\end{lemma}
\begin{defn}[Metric space]
    A metric space is a pair $(\bX, d)$ where $\bX$ is a set and $d:\bX\x\bX\rightarrow \R$ is a map such that for all $x, y, z\in\bX$ the following conditions hold:
    \[
        \begin{aligned}
            i.~~& d(x, y)\geq 0; & iii.~~&d(x,y) = d(y,x);\\
            ii.~~& d(x, y) = 0;~\Leftrightarrow x=y \quad & iv.~~&d(x,y)\leq d(x,z) + d(z, y).
        \end{aligned}
    \]
    The function $d$ is called ``metric''. Note that a metric space is called \textbf{compact} if $\bX$ is closed and bounded.
\end{defn}
\begin{defn}[Convergent sequence]
    Given a metric space $(\bX, d)$, a sequence $\{x_t\}_{t=0}^\infty$ is said to converge to some $x^*\in\bX$
    \[
        \forall \epsilon>0~~\exists t^*~~:~~\forall t>t^*~~d(x^*, x_t)<\epsilon.
    \]
\end{defn}
\begin{defn}[Cauchy sequence]
    A sequence $\{x_t\}_{t=0}^\infty$ is $\bX$ is a Cauchy sequence if 
    \[
        \forall\epsilon >0~~\exists t^*~~:~~\forall s,t>t^*~~d(x_s, x_t)<\epsilon.
    \]  
\end{defn}
\begin{defn}[Complete metric space]
    A metric space $(\bX,d)$ is complete if every Cauchy sequence in $\bX$ is convergent in $\bX$.
\end{defn}
\begin{defn}[Hausdorff space]
    Let $(\bX, d)$ be a complete metric space, and define $\cH(\bX)$ as the set of all compact subsets of $\bX$:
    \[
       \cH(\bX) = \{\bA\subset \bX : \bA~\text{is compact}\}.
    \]
\end{defn}
\begin{defn}[Hausdorff metric]
    Let $(\bX, d)$ be a metric space and let $\bY,\bZ\subset \bX$. The \textbf{Hausdorff metric} $d_\cH:\cH(\bX)\times\cH(\bX)\rightarrow \R$ is then defined as 
    \[
        d_\cH(\bY, \bZ) = \max \left\{\sup_{y\in\bY} d(y, \bZ), \sup_{z\in\bZ} d(\bY, z),\right\}
    \]
    where $d(y,\bZ) := \displaystyle\inf_{z\in\bZ} d(y,z)$.
\end{defn}
\begin{theorem}[Completeness of Hausdorff metric space \citep{barnsley1993fractal}]
    Let $(\bX, d)$ be a complete metric space. Then $(\cH(\bX), d_\cH)$ is a complete metric space. 
\end{theorem}
\citep{barnsley1993fractal} calls $(H(\bX), d_\cH)$ \textit{the space where fractals live}. It is the space where the mathematical foundations necessary to generate fractals via iterated function systems are developed.
\subsection{Contraction mappings}
\begin{defn}[Lipschitz function]
    Let $(\bX, d)$ be a metric space. A map $f:\bX\rightarrow\bX$ is \textit{Lipschitz} with constant $\ell$ if there exists $\ell>0$ such that
    \[
        \forall x, y\in\bX \quad d(f(x), f(y)) \leq \ell d(x, y).
    \]
\end{defn}
A Lipschitz function is then called \textit{contractive} iff $\ell<1$. Moreover, if $f:\bX\rightarrow\bX$ is Lipschitz, then $f$ is continuous.
\begin{tcolorbox}[enhanced, frame hidden, drop fuzzy shadow, colback=blue!5]
    {\textbf{Note:}} A map $f$ is contractive if brings any elements of $\bX$ \textit{close together}. The Lipschitz constant $\ell$ measures (bounds) how closer the points are brought together by one application of $f$. Thus, intuitively, if we iterate a discrete dynamical system 
    \[
        x_{t+1} = f(x_t)
    \]
    starting from any point of $x_0$ within $\bX$, the sequence $\{x_t\}_{t=0}^\infty$ will converge to a \textbf{unique} fixed point $x^*$ such that $f(x^*) = x^*$. 

    It is worth to notice that linear affine maps $f:\R^n\rightarrow\R^n$ (where we consider the standard Euclidian distance to form a complete metric space on $\R^n$)
    \[
        f(x) = Ax + b
    \]
    are contractive if and only if 
    \[
        \|A\|_2 = \sup_{x\in\R^n}\frac{\|Ax\|_2}{\|x\|_2}<1
    \]
    If this is the case any sequence $\{x_t\}_{t=0}^\infty$ would converge to a fixed point 
    \[
        x^* = (I - A)^{-1} b.
    \]
\end{tcolorbox}
The above intuitions can be formalized in the following classic result
\begin{theorem}[Banach fixed--point theorem]\label{th:bfpt}
    Let $(\bX, d)$ be a (non--empty) complete metric space and let $f:\bX\rightarrow\bX$ be a contractive map. Then $f$ admits a \textbf{unique} fixed point $x^*\in\bX$ such that any sequence $\{x_t\}_{t=0}^\infty$ defined by the iteration 
    \[
        x_{t+1} = f(x_t)
    \]
    converges to $x^*$ for any starting point $x_0\in\bX$, i.e.
    \[
        \exists!x^*\in\bX\quad\lim_{t\rightarrow\infty} x_t = \lim_{t\rightarrow\infty} f(x_t) = x^*\quad\forall x_0\in\bX.
    \]
\end{theorem}
\begin{cor}[Collage theorem]\label{cor:cot}
    Under the assumptions of Theorem \ref{th:bfpt}, it holds
    \[
        \forall x\in\bX\quad d(x, x^*) \leq \frac{1}{1-\ell} d(x, f(x)).
    \]
\end{cor}
\subsection{Iterated Function Systems}
With the aim of deriving fractal compression algorithms it is necessary to define functions on the Hausdorff metric space $(\cH(\bX), d_\cH)$. In particular, let $\{f_1, f_2,\dots, f_K\}$ be a collection of maps on $\bX$, $f_k:\bX\rightarrow\bX$. Then, we can define maps $F:\cH(\bX)\rightarrow\cH(\bX)$ by
\[
    F(\bA) = \bigcup_{k=1}^K f_k(\bA)\quad\forall\bA\in\cH(\bX)
\]
where $f_k(\bA)$ is intended as $f_k(\bA) = \{f_k(a) : a\in\bA\}$. Moreover, note that $\bA\in\cH(\bX)$ is also a compact subset of $\bX$. The following results shows that if all $f_k$ are contractive, then also $F$ is.  
\begin{theorem}[Contractivity of maps on the Hausdorff metric space]\label{th:ifs_contractivness}
    If for all $k=1,\dots, K$, the maps $f_k:\bX\rightarrow\bX$ are contractive with Lipschitz constant $\ell_k<1$, then $F:\cH(\bX)\rightarrow\cH(\bX); \bA\mapsto\bigcup_{k=1}^Kf_k(\bA)$ is contractive in the Hausdorff metric with Lipschitz constant $L = \max_{k}\{\ell_k\}_k$.
\end{theorem}
\begin{defn}[Iterated function system (IFS)]
    An iterated function system is a collection $\{f_1, \dots, f_K\}$ of contractive maps $f_k:\bX\rightarrow\bX$, represented by $F:\cH(\bX)\rightarrow\cH(\bX)$.
\end{defn}
Thanks to its contractivess (as established by Theorem~\ref{th:ifs_contractivness}), $F$ define a unique fixed point (\textit{attractor}) by the \textit{Banach fixed--point theorem} (Theorem~\ref{th:bfpt}), i.e. the discrete iteration defined by 
\[
    \bA_{t+1} = F(\bA_t) = \bigcup_{k=1}^K f_k(\bA_t)
\]
converges to $\bA^*\in\cH(\bX)$ for $t\rightarrow \infty$.
Since the attractor $\bA^*$ is unique, it is completely defined by the map $F$. The \textit{data encoding problem} can be then formulated as follows.
\begin{tcolorbox}[enhanced, frame hidden, drop fuzzy shadow, colback=red!5]
    {\textbf{Problem:}} \textit{Fractal Data Encoding} (Section 2.2)\\
    \begin{center}
        If we are given some set $\bS\in\cH(\bX)$ (our \textit{data}), can we find map $F$ whose attractor is $\bS$?
    \end{center}
    In other words, given data $\bS\in\cH(\bX)$, find the collection of maps $f_k:\bX\rightarrow\bS$ such that the following conditions hold
    \[
        \begin{aligned}
                i.~~& \text{$F:\cH(\bX)\rightarrow\cH(\bX); \bA\mapsto\bigcup_{k=1}^K f_k(\bA)$ is contractive};\\
                ii.~~& \text{$\bS$ is \underline{the} fixed point of $F$},~\bS = F(\bS) = \bigcup_{k=1}^K f_k(\bS);
        \end{aligned}
    \]
\end{tcolorbox}
\paragraph{Properties of data encoding} Note that condition $ii.$ suggests that, in order for the problem to admit a solution, data should be made up of transformed copies of itself. Specifically, we are assuming that it is possible to take the data $\bS$, copy it $K$-times, apply to the copies some contractive transformations and finally stitch them together to reconstruct the initial data $\bS$. The uniqueness of an attractor induced by the contractivity of $F$ is fundamental to practically solve the fractal encoding problem because if we can find an $F$ such that $\bS = F(\bS)$, then we will be sure that $F$ is the unique solution of the encoding problem.

\paragraph{On the Collage Representation} By applying the \textit{Collage Theorem} (Corollary~\ref{cor:cot}) to $F$ using the Hausdorff metric $d_\cH$ we have 
\[
    \begin{aligned}
        & d_\cH(\bS, \bA^*)\leq \frac{1}{1 - L}d_\cH(\bS, F(\bS))\\
        \Leftrightarrow~~ &  d_\cH(\bS, \bA^*)\leq \frac{1}{1 - \max_k\{\ell_k\}}d_\cH\left(\bS, \bigcup_{k=1}^Kf_k(\bS)\right)
    \end{aligned}
\]
This means that if we can't stitch the transformed copies $f_k(\bS)$ together to perfectly reconstruct the data $\bS$, i.e.
\[
    d_\cH\left(\bS, \bigcup_{k=1}^Kf_k(\bS)\right) \neq 0\quad(\Leftrightarrow \bS\neq F(\bS)),
\]
then the lower the Lipschitz constant $L$ of the IFS is, the lower the distance between the data $\bS$ and the attractor $\bA^*$ of $\bS$ will be given a mismatch $d_\cH(\bS,F(\bS))$. As mentioned in the main text, this implicitly promotes the use of ``very contractive'' maps $f_k$ (i.e. with low $\ell_k$).
\paragraph{A learning perspective to the \textit{fractal data encoding}} In the language of machine learning practice, the \textit{fractal data encoding} problem can be translated into finding a parametric representation $f_k(~\cdot~; w_k)$, $w\in\R^{n_w}$ for the functions $f_k(\cdot)$ (e.g. Neural Networks with parameters $w_k$) where the parameters $w = (w_1,\dots, w_K)\in\bW$ are trained to minimize the Hausdorff metric loss function $d_\cH (\bS, F(\bS; w))$ naturally induced by the \textit{Collage Theorem}, i.e.
\[
    \begin{aligned}
        \min_w~~& d_\cH (\bS, F(\bS; w))\\
        \text{subject to}~~& F(\bS; w) = \bigcup_{k=1}^K f_k(\bS; w_k)\\
        & w \in\bW
    \end{aligned} 
\]
Once optimal $f_k(~\cdot~; w_k)$ are computed, it is easy to find the data that $F$ encodes (i.e. the \textit{decoding} process): after sampling any initial condition $\bA_0$, the encoded data can be obtained by iterating 
\[
    \bA_{t+1} = F(\bA_t),
\]
until convergence to $\bA^*\approx \bS$.
\subsection{Partitioned Iterated Function Systems}
Existance of solutions of the fractal encoding (inverse) problem requires data to be perfectly representable by an IFS. Conversely, we can define \textit{self-similar} data if it is the attractor of an IFS. 
\begin{defn}[Self--similar sets]
    A set $\bS$ is called self--similar if and only if there exists a contractive map $F:\cH(\bX)\rightarrow \cH(\bX)$ whose attractor is $\bS$, i.e. $\bS = F(\bS)$.
\end{defn}
While verifying the self--similarity of a specific data point is undoubtedly a NP--hard problem, natural data (e.g. in an image dataset) is unlikely to satisfy this strict property. The challenge is that the self-similarity property has to be global across the set $\bS$. That is, the entire set $\bS$ has to be made up of smaller copies of itself, or parts of itself. If one zooms in on it, it would display the same level of detail, regardless of the resolution scale \citep{welstead1999fractal}.

For this reason, it is necessary to extend the fractal encoding to more general, non globally self--similar sets. This can be achieved by introducing the technology of \textit{partitioned function systems} (PFS) where the domains of the contraction maps $f_k$ are restricted. 
\begin{defn}[Partitioned Function System]
    Let $(\bX, d)$ be a complete metric space and let $\bD_k\subset \bX$ for $k=1,\dots,K$. A partitioned function system is a collection of contraction maps $f_k:\bD_k\rightarrow\bX$.
\end{defn}
Note that, according to \cite{fisher2012fractal}, it is not possible to extend  Theorem \ref{th:bfpt} to PFSs in the general case to effectively ensure existance and uniqueness of fixed points. Intuitively this is due to the fact that the domains of $f_k$ are restricted and the convergence of the decoding dynamics (i.e. the fixed-point iteration)
\[
    \bA_{t+1} = \bigcup_{k=1}^K f_k(\bA_t)
\]
becomes dependent on the choice of the initialization $\bA_0$. In fact, even though if choose $\bA_0\subset\bigcap_{k=1}^K \bD_k$, after one step we may end up with an empty set. Note that this is generally not a problem in practice when applying PFSs to or $\CO$ working with common type of data such as images or audio signals.
\subsection{Functional Representation of Data}
In order to derive an implementation--oriented formulation of data encoding with partitioned functions systems in the general case, it can be convenient to rely on a \textit{functional} description of data (see e.g. \citep{welstead1999fractal, fisher2012fractal}). In example, images (of infinite resolution) can be represented as functions from the unit square to $\R$. Time series can also be thought as real continuous functions over a compact time domain.

Specifically we restrict our analysis to the space $\cF = \{\phi: \dom(\phi)\rightarrow\R\}$ of \textit{data} defined the \textit{graphs} $(z, \phi(z))$, $z\in\dom(\phi)$ of (measurable) functions over the compact domain $\dom(\phi)$ and values in $\R$. $\dom(\phi)$ is assumed to be a compact subset of $\R^n$.

\paragraph{Partitioned fractal encoding \textit{a la} \cite{welstead1999fractal}}
We choose $\cF = L^p(\cX;\R)$ with $\cX$ a compact subset of $\R^n$ and we equip it with a metric $d_\cF$ induced by the Lebesgue measure
\[
    d_\cF(\phi, \psi) = \left(\int_\cX |\phi(x) - \psi(x)|^p\dd x\right)^{1/p}.
\]
Then $(\cF, d_\cF)$ is a complete metric space and the \textit{Banach fixed--point} (Theorem~\ref{th:bfpt}) holds. Then, we specialize the partitioned function system on $(\cF, d_\cF)$ as comprised of the following collections of $K$ elements:
\[
    \begin{aligned}
          a.~~ & \textit{sub--domains}~\sD_k\subset\cX;\\
          b.~~ & \textit{invertible contractive maps}~v_k: \sD_k\rightarrow\sR_k\subset\cX;\\
          c.~~ & \textit{maps $f_k:\cF\rightarrow\cF$ defined as}\\
               &\quad f_k(\phi)(x) = c_k \phi(v^{-1}_k(x)) + d_k\quad \forall \phi\in\cF,x\in\cX.
    \end{aligned}
\]
Note that we can define subsets $\sR_k$ as the range of $v_k$ operating on $\sD_k$, i.e. $\sR_k = v_k(\sD_k)$. The constants $c_k, d_k$ realize an affine trasformation on $\phi$ by expanding/contracting and shifting the range of $\phi$. The contractive maps $v_k$ are the ``spatial part'' of the PFS and map the domains $\sD_k$ to their respective ranges $\sR_k$. $v_k$ are often chosen to be affine maps
\[
    v_k(x) = A_k x + b_k, \quad A_k\in\R^{n\times n},b_k\in\R^n.
\]
Note that it is possible to choose $A_k$ and $c_k$ so that $f_k$ is contractive. In particular, it is sufficient to require $|c_k||\det A_k|^{1/p}<1$
\begin{defn}[Tiling partition]
    A collection of ranges $\sR_k\subset\cX$ is said to tile $\cX$ iff $\cX = \bigcup_{k=1}^K\sR_k$ and $\forall i\neq j~~ \sR_i\cap\sR_j = \emptyset$.
\end{defn}
If the ranges $\sR_k$ \textit{tile} $\cX$, we can define the operator $F:\cF\rightarrow\cF$ by
\[
    F(\phi)(x) = f_k(\phi)(x)~~\text{for}~~x\in\sR_k,
\]
i.e.
\[
    \begin{aligned}
        F(\phi)(\cX) &= \bigcup_{k=1}^K f_k(\phi)(\sR_k)
        = \bigcup_{k=1}^K c_k\phi(\sR_k) + d_k
        = \bigcup_{k=1}^K c_k\phi(v_k^{-1}(\sD_k)) + d_k
    \end{aligned}   
\]
Since the ranges $\sR_k$ tile $\cX$, $F$ is defined for all $x\in\cX$, so $F(\phi)$ is a function of the same class of $\phi$. 

If $\phi$ is an image on the unit square tiled by the ranges $\sR_k$, then $F(\phi)$ will also be an image on the unit square.

Assuming all the maps $f_k$ to be contractions on $\cF$, $F$ satisfies the Banach fixed point theorem and has unique fixed point $\phi^*\in\cF$ such that 
\[
    \begin{aligned}
        \phi^* = F(\phi^*).
    \end{aligned} 
\]

\paragraph{Collage and PIFS for digital images}
Similarly to the main text, we can define PIFS by restricting our analysis to \textit{affine} maps, operating on the space of discrete images of a given resolution with a total number $m$ of pixels. Note that pixels across channels can be treated effectively as different elements.

We assume the value of each pixel to range in $\R$ and to collect all the pixel values in an \textit{ordered}\footnote{with a specific predefined criterion.} in a vector $z\in\R^m$. Then, a \textit{partitioned function system} \citep{jacquin1993fractal, welstead1999fractal, fisher2012fractal} can be represented as the \textit{structured} map
\begin{defn}[Discrete PIFS]
    Consider a $m$-pixel image represented by the ordered vector $z\in\R^m$. Then, a \textit{Discrete PIFS} is defined as the parametric linear map $F:\R^{m}\rightarrow\R^m$:
   \begin{equation}\label{eq:dco}
       F(z; w) = \sum_{k=1}^K {\color{blue!70}a_k} T_k P_k S_k z + \sum_{k=1}^k {\color{blue!70}b_k} T_k \1,\quad w = (a_1,\dots,a_K,b_1,\dots b_K)
   \end{equation}
    where $T_k,~P_k, S_k$ are defined similarly to Definition~\ref{def:nco}.
\end{defn}
The output of the collage operator for $\sR_k$ is thus a pooled and scaled version of $\sD_k$ translated block--wise by $b_k$. A symbolic formulation of the collage operator can be also given by
\[
    \sR_k = f_k(\sD_k; w_k),\quad w_k = (a_k, b_k)
\]
Since the collection of range cells $\sR_k$ tiles the whole image $\sI$, we can write (\textit{a la} IFS)
\[
    F(\sI; w) = \bigcup_{k=1}^K f_k(\sD_k; w_k)
\]

\begin{remark}[Extensive search]
    Note that, in the classic setting of \cite{jacquin1993fractal, welstead1999fractal}, for each range cell $\sR_k$, the corresponding domain cell $\sD_k$ has to be found by \textit{extensive search} through the set of all possible pooled domain cells. 
\end{remark}
We provide a compact algorithmic summary of the core steps in fractal compression as per \cite{jacquin1992image,jacquin1993fractal} in Figure \ref{fig:algo_fc}. Other variants of fractal compressions have historically been attempted, including ones with adaptive partitions and different algorithms to solve the combinatorial search during encoding.
\begin{figure}
\begin{algorithm}[H]
\centering
\caption{Fractal Compression with PIFS \cite{jacquin1992image}\\\fcircle[fill=wine]{2pt} ~Encoding~ \fcircle[fill=deblue]{2pt}~Decoding}
    \begin{algorithmic}
        \STATE \textbf{Input:} Image $\sI$, parametrized PIFS $F(~\cdot~; w)$.
        \STATE Partition $\sI$ into two collections of $N$ domains $\sD_n$ and $K$ ranges $\sR_k$.
        \FOR{$k$ \textbf{from} $1$ \textbf{to} $K$}
            \STATE $\sD_n^*, w_k^* = \displaystyle\arg\min_{\sD_n,w_k}  d(f_k(\sD_n; w_k), \sR_k)$ \hfill \textit{matching}
        \ENDFOR
        \STATE Store ${\tt code}:=\{w^*_k,  (n, k)\}_{k=1}^{K}$ \hfill \textit{fractal code and domain index}
        \STATE Initialize $\tilde \sI_0$ as any image with same size as $\sI$
        \REPEAT 
        \STATE $\displaystyle\tilde I_{t+1} = \bigcup_{k=1}^K f_k(\sD_{n, t}^*; w_k^*)$
        \UNTIL{convergence}
        \RETURN $\tilde I^*$ \hfill \textit{decoded data}
    \end{algorithmic}
\end{algorithm}
\vspace{-6mm}
\caption{Algorithmic summary of fractal compression with PIFS.}
\label{fig:algo_fc}
\end{figure}

\clearpage

%% file: appendix_arxiv/C_additional.tex
\section{Additional Details}
\paragraph{Code--length flexibility: $\CO$ and PIFS}
A $\CO$ operator is a generalization of PIFS operators of fractal compression algorithms \citep{jacquin1992image,jacquin1993fractal,fisher2012fractal}. In particular, the $\CO$ introduces additional flexbility in the choice of \textit{code length} i.e. how many bits to allocate to the compression code. For example, given non--adaptive (square) tiling domain and range cells\footnote{Although different choices are possible, "tiling" partitions are most convenient; when applying $F$, a domain partition $\sD$ can be identified via single  $\lceil{\log_2{N}}\rceil$--bit integer address.}, the \textit{bits--per--dimension} (bpp) cost of PIFS--based fractal compression of \cite{jacquin1992image} is $\frac{{\tt cost}_{w_k}}{n_{r,k}^2}$, where ${\tt cost}_{w_k}$ is the cost of saving the parameters of single element in the $f_k$ of $F_w$. 

Here, other than seeking further compression and reduction of ${\tt cost}_{w_k}$ by modifying the class of operator, the only degree--of--freedom is to reduce or increase the dimensions of tiling partitions. Note that modifying the partition scheme has not only effect on the bpp cost but also on the type of self--similarity that can be captured. Instead, $\CO$ operators offer an additional design axis; indeed, the bpp budget -- given a fixed partition -- can be modified by increasing or decreasing the number of auxiliary domains introduced as learnable feature maps. This number is independent on the number of original domains, whereas the number of additional domains generated as affine augmentations of fractal compression (see \cite{welstead1999fractal,fisher2012fractal} for details) is not. 

\paragraph{On adaptive partitions}
Elaborated partitioning schemes have been developed for PIFS--based fractal compression methods \citep{fisher2012fractal}. While the analysis and empirical comparisons of this work have been centered around the operators, rather than partition schemes, we remark that $\CO$ are compatible with alternative and potentially adaptive schemes. Much like for PIFS, this is a likely direction for further improvement of Neural $\CO$.

\subsection{Extended related work}
\paragraph{Attention operators and patches}
There exist superficial similarities between $\COs$ and attention operators \citep{vaswani2017attention}. In particular, recent variants of vision transformers \citep{dosovitskiy2020image} where attention acts on square patches, can be seen as a single step of a $\CO$, where source and target partition match and aggregation weights are found via similarity scores. $\COs$ differ from attention in that they are structured fixed--point iterations, are built to accommodate non--overlapping partitions, are resolution--invariant, and have a compact parametrization that can be used as a compression code. It remains to be seen whether investigating attention operators through the lenses of $\COs$ can yield improvements in theoretical understanding or performance.

\paragraph{Fractal compression}
\cite{sun2001neural} parametrize elements of the iterative map with small neural networks. The proposed method still requires training on each image, with marginal improvements over standard variants. \citep{guido2006fractal} provide a preliminary exploration of fractal coding for audio. Despite the extensive body of work, fractal methods for image compression are rarely used in place of other codecs due to slow encoding. As discussed in the main text sections, Neural $\COs$ address this limitation via neural network amortization. We note that Neural $\CO$ remain compatible with adaptive partitioning schemes, which provides a likely avenue of further improvement. We highlight a line of work on different probabilistic models of self--similarity \citep{zha2020image} for tasks such as image restoration. 

\clearpage

%% file: appendix_arxiv/D_empirics.tex
\section{Additional Experiment Details}
\paragraph{Hardware and software}
The experiments have been performed on a workstation with $2$ \textsc{NVIDIA GeForce RTX 3090} GPUs. We use JAX\footnote{https://github.com/google/jax} for model implementation and distributed training. The code is available at \url{github.com/ermongroup/self-similarity-prior}.

\subsection{Neural $\COs$ for Fractal Art}
We construct the encoder $\cE$ for $w$ by stacking $4$ blocks composed of interleaved depthwise and pointwise convolutions. We train for $5000$ iterations on each image displayed in Figure \ref{fig:art} with AdamW \citep{loshchilov2017decoupled}. We produce $\sU$ by augmenting at each step of the $\CO$ with rotations of $[90, 180, 270]$ degrees and flips, produced by multiplying all pixels values of a domain cell by $-1$. We do not use any additional learned auxiliary domain, so that the patterns can be kept globally fractal. 

\paragraph{Neural $\CO$ Texturizers} We report additional results in \ref{texturization}, where a Neural $\CO$ is used to texturize images by optimizing transformations of a fixed $\sU$, provided as external "texture source". To promote utilization of texture sources we introduce a coefficient to weigh $\sU$ relative to $\sD$ in the $\CO$ iteration. We note that in this case the Neural $\CO$ is not leveraging any self--similarity; rather, this should be intended as a display of the capability of a Neural $\CO$ to aggregate both external as well as self--referential information to achieve a given task.

\subsection{Neural $\COs$ for Generation}
We design the architecture of a $\CO$ following \citep{child2020very}. Table \ref{tab3:model_hparams} describes the model structure. We introduce $30$ learned auxiliary domains, parametrized to be pixel patches of same size of range cells. As the tiling partition, we choose a single domain cell of size $28\times 28$ and size $4$ range cells of $14\times14$. All models use a Bernoulli likelihood. Note that in this case, the fixed--point of the generator $p(x|w)$, chosen as a $\CO$, is given by the collection of all Bernoulli parameters, one for each pixel. This is thus an example of a $\CO$ that is does not decode pixel--values of an image as its fixed--point, but rather parameters of their distributions. We optimize the ${\tt ELBO}$ by sweeping the KL weight $\beta$ as discussed in Figure \ref{fig:samples} for $2000$ epochs. Additional training details are provided in \ref{tab3:model_hparams2}.

\subsection{Neural $\COs$ for Compression}
We construct the encoder $\cE$ for $w$ by stacking $4$ blocks composed of interleaved depthwise and pointwise convolutions. We train for $5$ epochs with AdamW \cite{loshchilov2017decoupled} on a dataset of $8000$ crops of size $40\times 40$ obtained from the DOTA \cite{xia2018dota} aereal image training dataset. The dataset is generated (statically) randomly by applying a random rotation, followed by a random crop. We produce the $10$ held--out images of size $1200\times1200$ with a similar procedure, applied to the test dataset. We note that DOTA images are all of different resolutions, motivating the above procedure. Speedup results are provided in Figure \ref{fig:speedups}. 

As baselines, we use the official COIN \cite{dupont2021coin} implementation. We develop a GPU--parallel version of fractal compression with PIFS \cite{jacquin1992image,jacquin1993fractal,welstead1999fractal,fisher2012fractal} as a baseline. We use the same partition strategy as for Neural $\COs$, namely tiling into domain and range cells. Our evaluation includes a fractal compression variant which incorporate $\sU$ by augmenting domains, $\sD$ at each step, with rotations of $[90, 180, 270]$ degrees and flips, produced by multiplying all pixels values of a domain cell by $-1$. The matching problem of domains to ranges is solved via least--squares as per \cite{welstead1999fractal}. We parallelize the least--square solving across domains.  

\subsection{Computation of bits--per--pixel}
We report the per--image \textit{bits--per--pixel} (bpp) cost of compression baselines and Neural $\CO$ compressors.

\paragraph{Neural $\CO$} Computation of the bpp of fractal codes generated by a Neural $\CO$ compressor requires the following considerations. First, mixing weights $\gamma_{k,n}$ are premultiplied to both $a_{k,n}$ and $b_{k,n}$. The same holds for mixing weights of auxiliary domains. We store a single $b_k$ for each $\sR_k$, noting that the corresponding term can be precomputed as $b_k=\sum_{n=1}^{N} \gamma_{k,n} b_{k,n}$. Further, we exploit the a priori knowledge that $a,b\in[-1,1]$, enforced via {\tt tanh}, in combination with significant digit clipping. Quantizing by clipping to a threshold $\epsilon$ of significant digits, in combination with the bounds enforced by {\tt tanh}, allows bit--packing each into less than $\lceil\log_2{10^\epsilon}\rceil+2$. One of the $2$ additional bits is for sign information. This can be verified by noticing that by quantizing the range of values, after multiplying by $10^{\epsilon}$, is contained by the integer range $[-10^{\epsilon},10^{\epsilon}-1]$. In practice, the number of bits is less than $\lceil\log_2{10^\epsilon}\rceil+2$ since not all values in entire integer interval defined by $\epsilon$--quantization are utilized by $a_{n,k}$ and $b_k$ for a given image. In particular, we consider maximum absolute values of the quantized values, and restrict the interval accordingly.

Neural $\CO$ do not require storing of domain cell addresses, since each map of the $\CO$ corresponding to a range cell (see \ref{eq:step}) always transforms all domains. The specification of patch--sizes, especially when they are the same across all domains, and across all ranges, as well as type of pooling operators can be considered part of the codec, adding a negligible amount of bits. Further considerations are necessary in one wishes to employ more elaborate partition schemes \citep{fisher2012fractal}. 

The overall cost is given by
\begin{equation}
    {\tt bpp} = K(N+V){\tt bpp}_a + K{\tt bpp}_b + V{\tt bpp}_u
\end{equation}
where $N$ is the number of domains, $V$ is the number of learned auxiliary cells, and $K$ the number of ranges. We indicate with ${\tt bpp}_u$ the cost of saving auxiliary learned patches. This cost is amortized across each image of the held--out set, as the patches are the same for a given Neural $\CO$. For $V$ auxiliary patches of size $h\times w \times c$, the (non--amortized) bit cost is $32\cdot V\cdot h\cdot w\cdot c$ bits. We provide some example calculations in \ref{tab:compute_bpp}.

\begin{table*}[t]
    \centering
    \begin{tabular}{c|cccccccc}
                  & $K$ & $N$ & $V$ & $\epsilon$ & ${\tt bpp}_u$& ${\tt bpp}_a$& ${\tt bpp}_b$ & total \\ \midrule 
        low--bpp & $4$ & $1$ & $3$ & $3$ & $9 \cdot 10^{-4}$ & $6.6\cdot 10^{-3}$ & $6.3\cdot 10^{-3}$ & $0.134$ \\
        medium--bpp & $4$ & $1$ & $10$ & $4$ & $8.9\cdot 10^{-4}$ & $6.5\cdot 10^{-3}$ & $5.9\cdot 10^{-3}$ & $0.319$ \\
    \end{tabular}
    \caption{Example \textit{bits--per--pixel} (bpp) code length computation for Neural $\CO$. To determine ${\tt bpp}_a$ and ${\tt bpp}_b$ we consider respective maximum values and represent their quantized integer range as discussed in C.4. The cost of auxiliary inputs is amortized on the $10$ held--out $1200\times 1200$ crops of DOTA, as decoding uses the same learned auxiliary domains for all images. We consider the auxiliary cost ${\tt bpp}_u$ as part of the fractal code for a worst--case comparison, noting that reutilization of the same compressor eventually amortizes the cost to $0$. For $M$ images, $\lim_{m\to\infty}{\tt bpp}_u = 0$.}
    \label{tab:compute_bpp}
\end{table*}

\paragraph{COIN} We use the official implementation of \citep{dupont2021coin}, where the bpp is computed by serializing the weights of the network into a bytes. 

\paragraph{Fractal compression baseline} We quantize fractal compression affine maps $f_k$ into half--floats, $16$ bits for each $a_k$ and $b_k$. As the address of the domain cell associated to a given range $\sR_k$, we store the index with cost $\lceil{\log_2{(N+V)}}\rceil$--bit, where $N$ is the number of source domains $\sD$ and $V$ the number of auxiliary domains $\sU$. 

\paragraph{block--DCT} 

We apply a forward, two--dimensional \textit{discrete cosine transform} (DCT) to patches of sizes $12$ (high bpp) and $16$ (medium bpp) and filter all but the lowest coefficient. The total cost is thus $32 \cdot n_{\text{patches}}$. The image is decoded by applying an inverse DCT.

\clearpage

\begin{table*}
\centering
\setlength{\tabcolsep}{0.8em}
\resizebox{1\textwidth}{!}{
\begin{tabular}{@{} l c c c c c c c c @{}}\toprule
  & \multicolumn{2}{c}{\textbf{\textsc{Architecture}}} &
  & \multicolumn{4}{c}{\textbf{\textsc{Learning Parameters}}} \\
\cmidrule(lr){2-4} \cmidrule(l){5-8} 
Model & \sc{Encoder} & \sc{Decoder} & \sc{Channels}
& & \sc{$\beta$} & \sc{Decoder Latent} &\sc{Num. Auxiliary} \\
\cmidrule(r){1-1}\cmidrule(lr){2-4} \cmidrule(l){5-8} 
$\CO$ VAE & ``28x1,28d4,7x1,7d7,1x1'' & ``1x4'' & 128 & & [$0.5, 0.7, 1.0, 1.2, 1.5$] & 64 & [$30$] \\
VDVAE & ``$28$x$1$,$28$d$4$,$7$x$1$,$7$d$7$,$1$x$1$'' & ``$1$x$1$,$7$m$1$,$7$x$2$,$28$m$7$'' & $64$ & & [$0.5, 0.7, 1.0, 1.2, 1.5$] & $16$ & -- \\
\bottomrule
\end{tabular}
}
\caption{Autoencoder model setup for VDVAE and VDCVAE (ours). Both were trained on BMNIST with Bernoulli likelihood. The encoder and decoder architectures strings can be interpreted as follows: "$28$x$1$" indicates $1$ block of $28$ residual layers and "$28$m$7$" is the mixing a $7$ by $7$ activation into an upsampled $28$ by $28$ embedding. A block consists of standard autoencoder parameterization with a learned prior, posterior and latent projection output layer. The channel dimensions apply to both the widths of encoder and decoder blocks with the collage decoder variants having learned separate maps per channel. $\beta$ describes the KL-coefficient weighting in the ELBO objective.}
\label{tab3:model_hparams}
\end{table*}

\begin{table*}
\centering
\setlength{\tabcolsep}{0.8em}
\resizebox{1\textwidth}{!}{
\begin{tabular}{@{} l c c c c c c c c c @{}}\toprule
  & \multicolumn{3}{c}{\textbf{\textsc{Training}}} &
  & \multicolumn{4}{c}{\textbf{\textsc{Optimizer}}} \\
\cmidrule(lr){2-5} \cmidrule(l){6-9} 
Model & \sc{Batch Size} & \sc{Epochs} & \sc{EMA} &\sc{Per Step (secs)}
& & \sc{Learning Rate} & \sc{Weight Decay} &\sc{Optimizer} \\
\cmidrule(r){1-1}\cmidrule(lr){2-5} \cmidrule(l){6-9} 
$\CO$ VAE (ours) & $32$ & $2000$ & $0.$ & $1.7$ & & $1e-4$ & $0.$ & AdamW($0.9$, $0.9$) \\ 
VDVAE & $32$ & $2000$ & $0.$ & $2.2$ & & $1e-4$ & $0.$ & AdamW($0.9$, $0.9$) \\ 
\bottomrule
\end{tabular}
}
\caption{Optimization settings for training baseline VDVAE and $\CO$ VAE (ours) on dynamically binarized MNIST.}
\label{tab3:model_hparams2}
\end{table*}

%% file: appendix_arxiv/E_genmo.tex
\section{Additional Results}

\subsection{Super--Resolution of $\CO$ VAE Samples}

\begin{figure}[H]%
    \centering
    \includegraphics[scale=0.95]{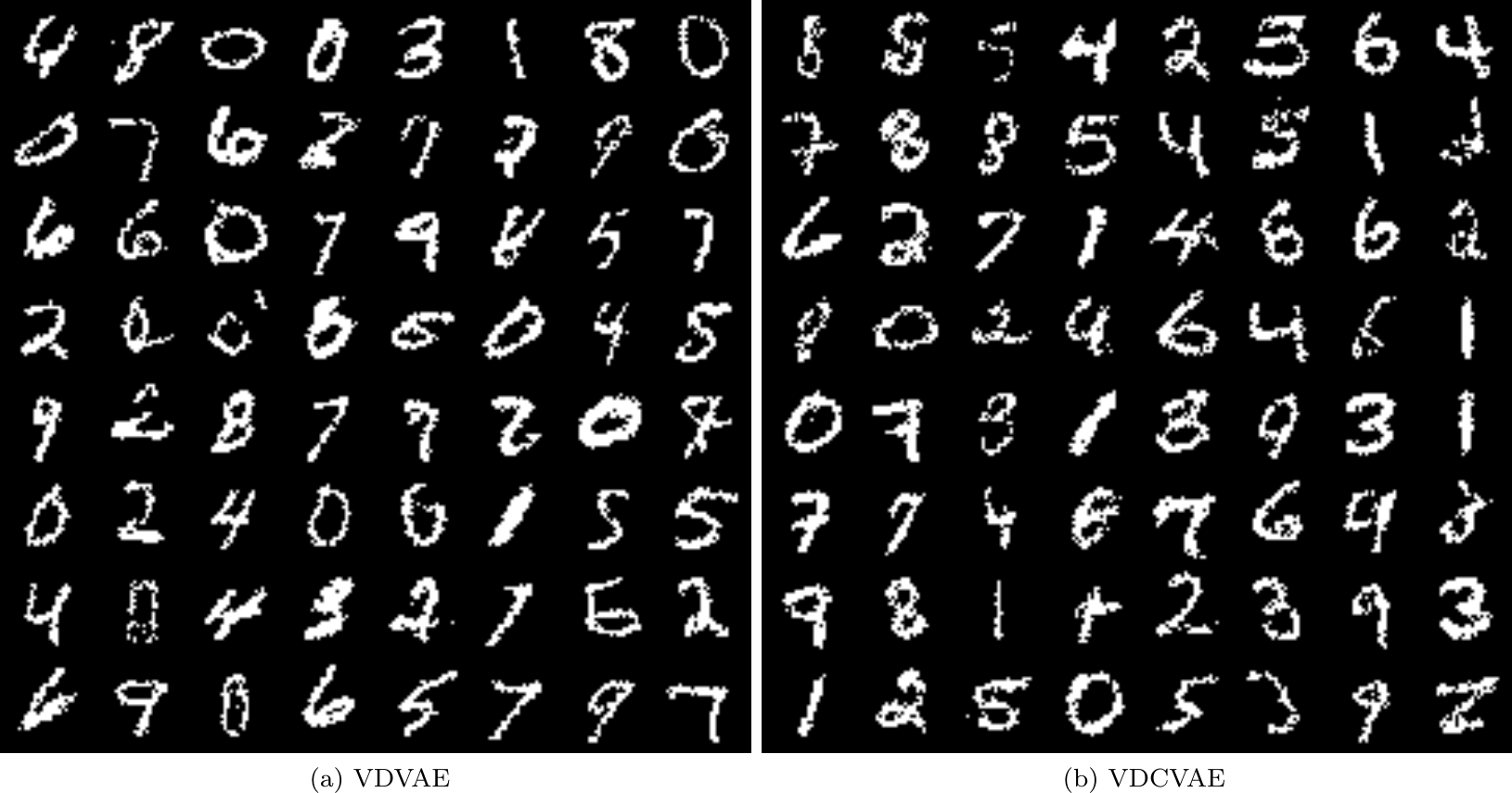}
    \caption{Super-resolution factor of $1\times$ the original data resolution ($28\times 28$ MNIST). \textbf{[Left]:} Samples from the VDVAE baseline. \textbf{[Right]:} Samples from a $\CO$ VAE.}%
    \label{fig:superresx1}%
\end{figure}

\begin{figure}[H]%
    \centering
    \includegraphics[width=0.49\linewidth]{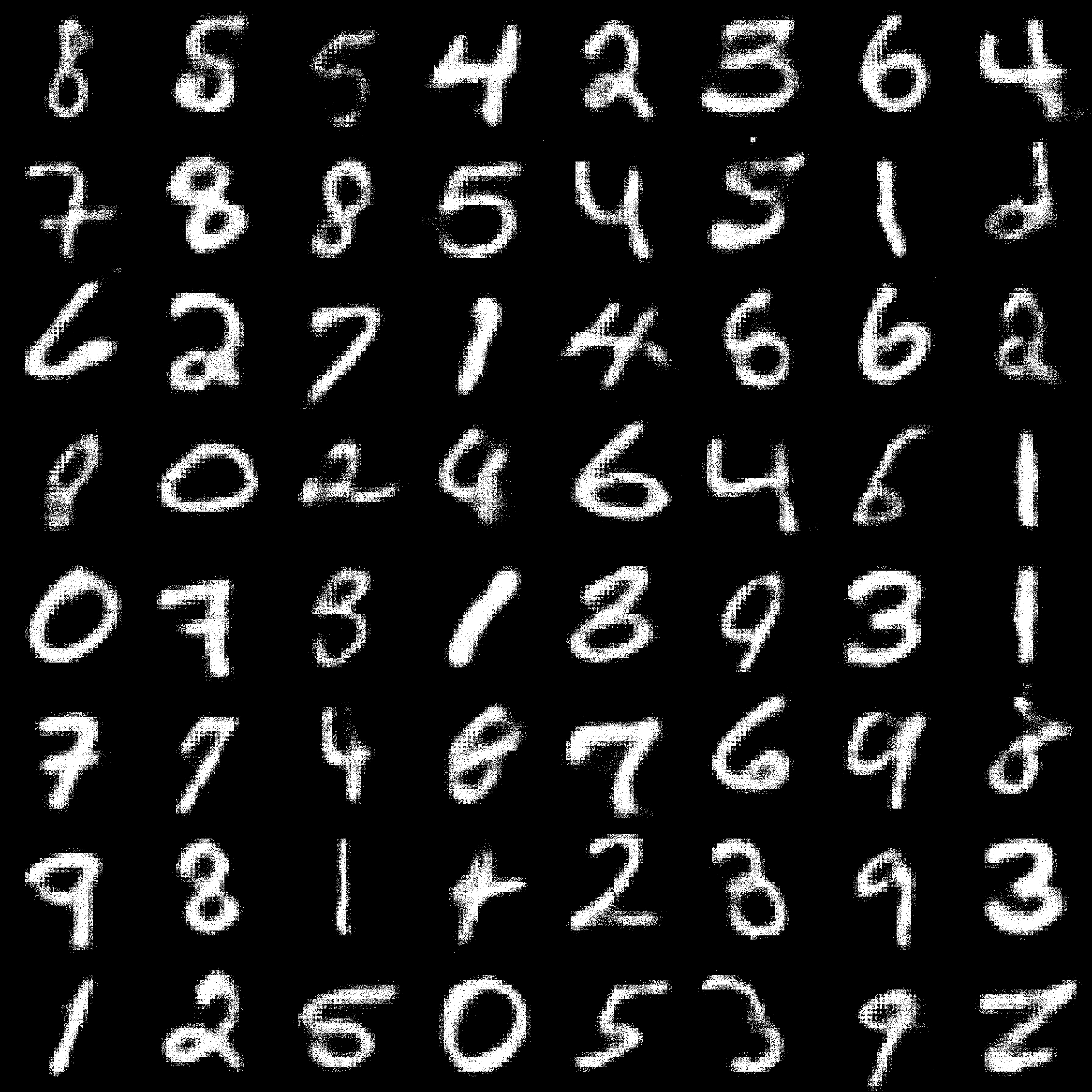}
    \includegraphics[width=0.49\linewidth]{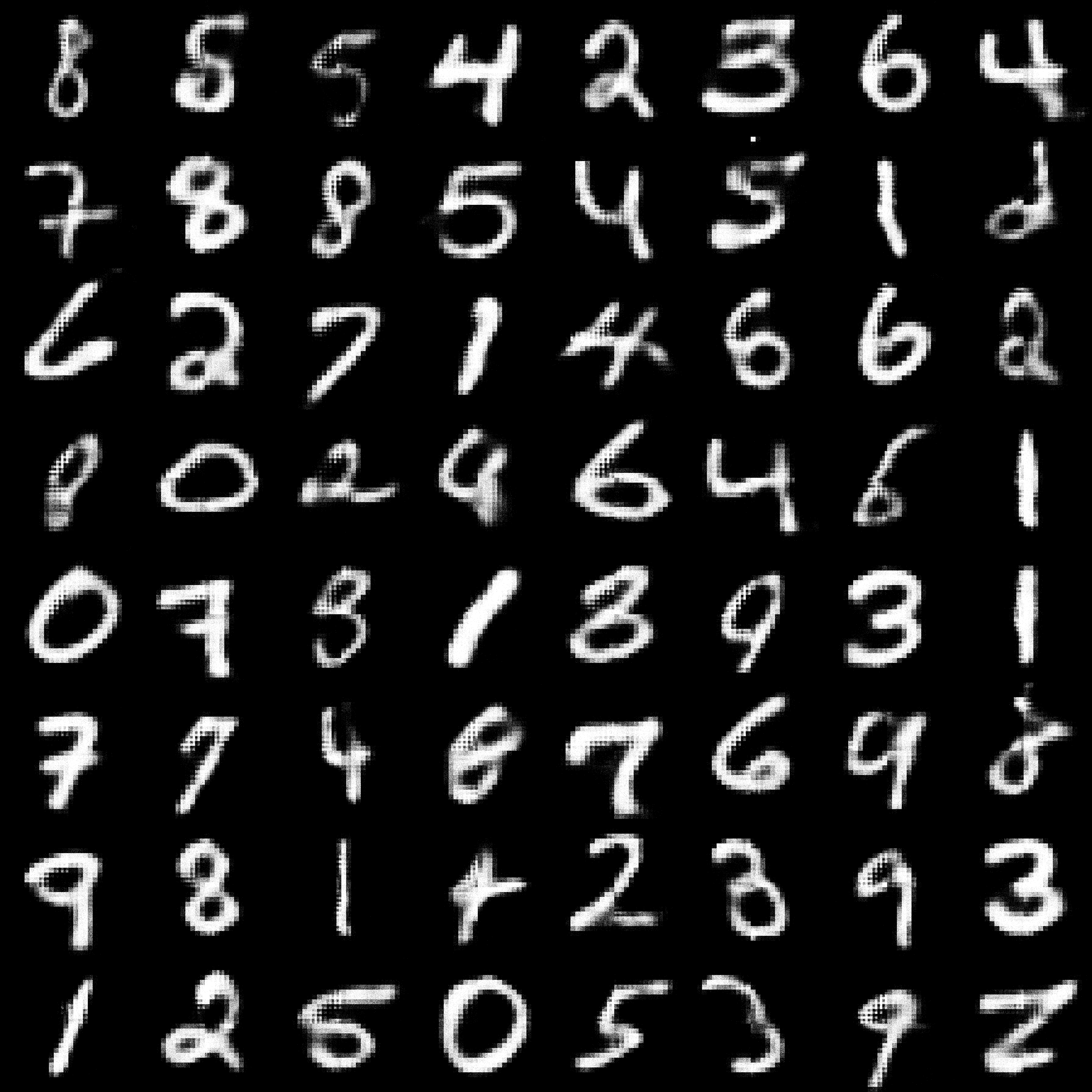}
    \caption{\textbf{[Left:]} $\CO$ VAE samples with $10\times$ magnification ($280\times 280$ resolution). \textbf{[Right:]} $\CO$ VAE samples with $40\times$ magnification ($1120\times 1120$ resolution).}%
    \label{fig:superresx40}%
\end{figure}

\subsection{Compression}
\begin{figure}[H]
\centering
\includegraphics[scale=1.]{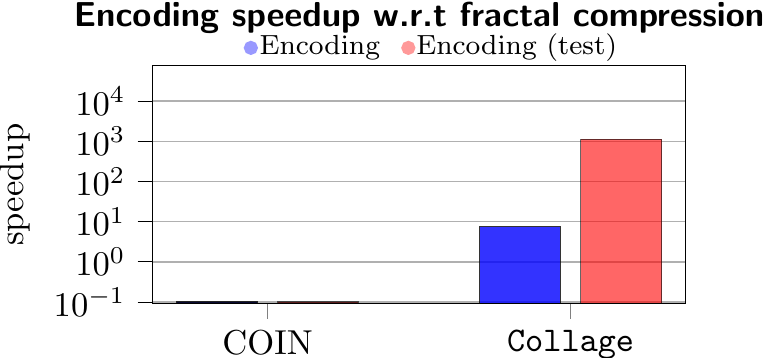}
\vspace{-1mm}
\caption{Wall--clock encoding speedups of Neural $\CO$ compressors and COIN over a PIFS--based fractal compression implementation on GPU.}
\label{fig:speedups}
\end{figure}

\subsection{Fractal Stylization}\label{texturization}

\begin{figure}[t]%
    \centering
    \includegraphics[scale=0.95]{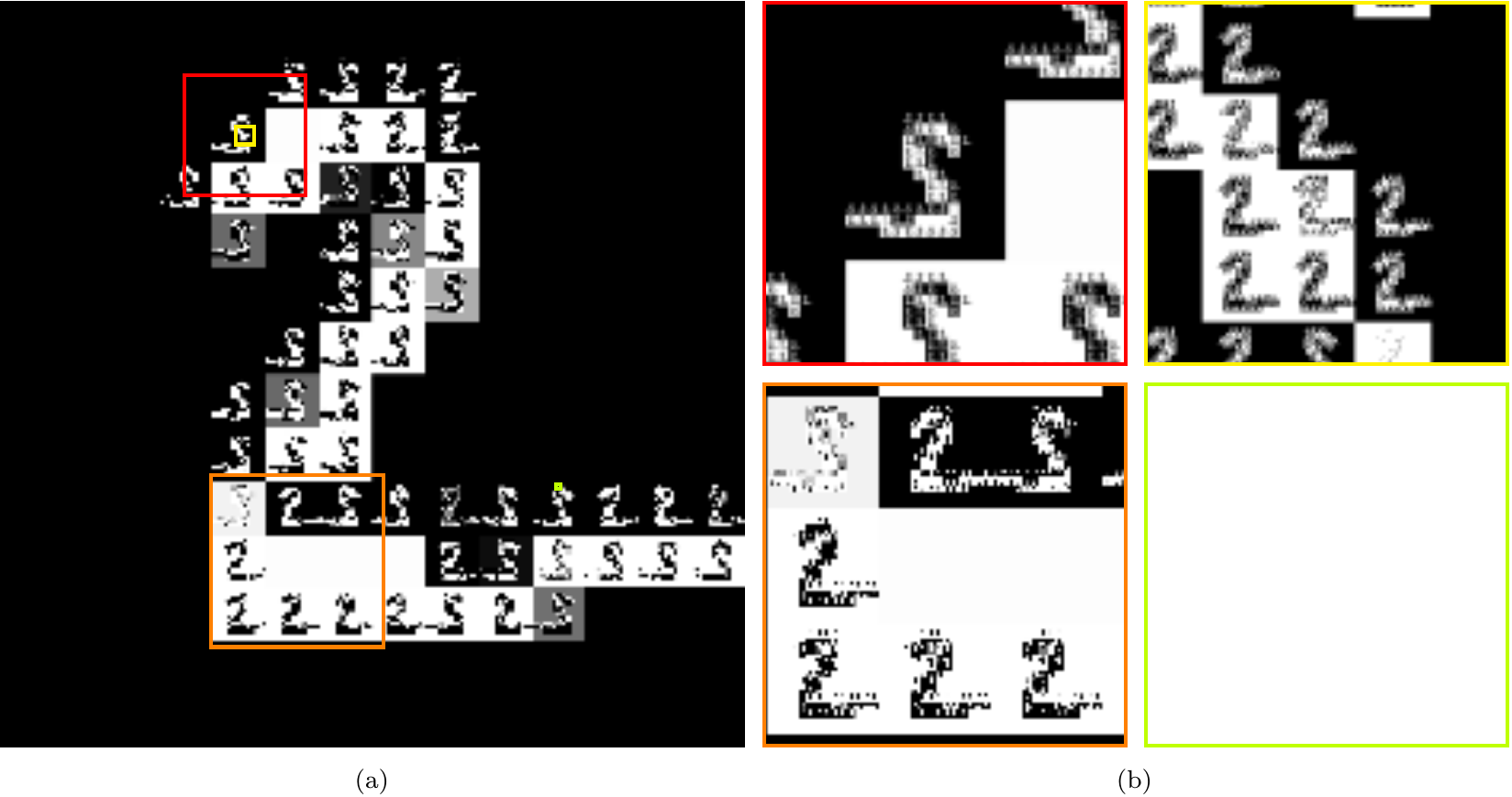}
    \caption{Fractalized MNIST digit via a \texttt{Collage} at $500\times$ the original data resolution. Red box is a $3\times$ magnification, yellow is a $20\times$ magnification, orange is $2\times$ magnification, and lime is $80\times$ magnification. Note that the image is compressed for easier displaying within the pdf. Refer to the code repository for the full quality version.}%
    \label{fig:mnist_fractalize_500x}%
\end{figure}

\begin{figure}[t]
\centering
\subfloat[\centering]{{\includegraphics[width=0.25\textwidth]{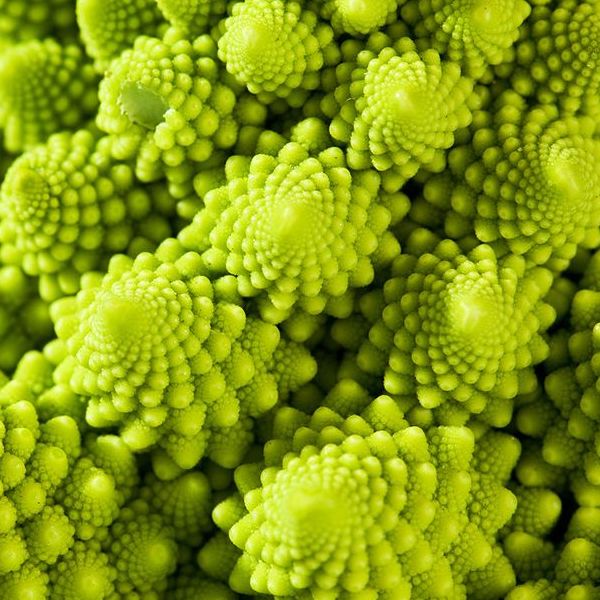} }}
\subfloat[\centering]{{\includegraphics[width=0.5\textwidth]{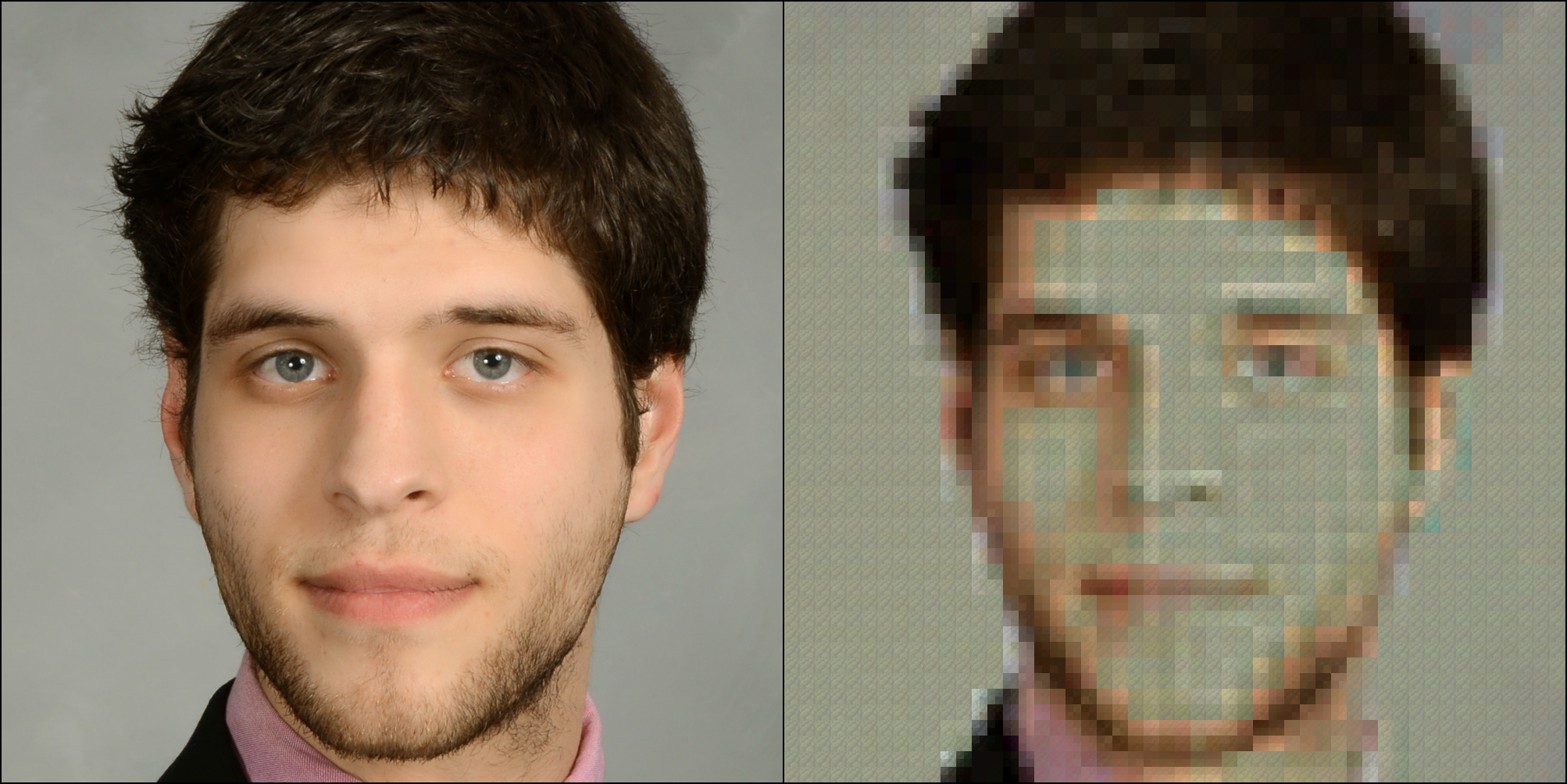} }} \\
\subfloat[\centering]{{\includegraphics[width=0.25\textwidth]{figures/romanesco.jpg} }}
\subfloat[\centering]{{\includegraphics[width=0.5\textwidth]{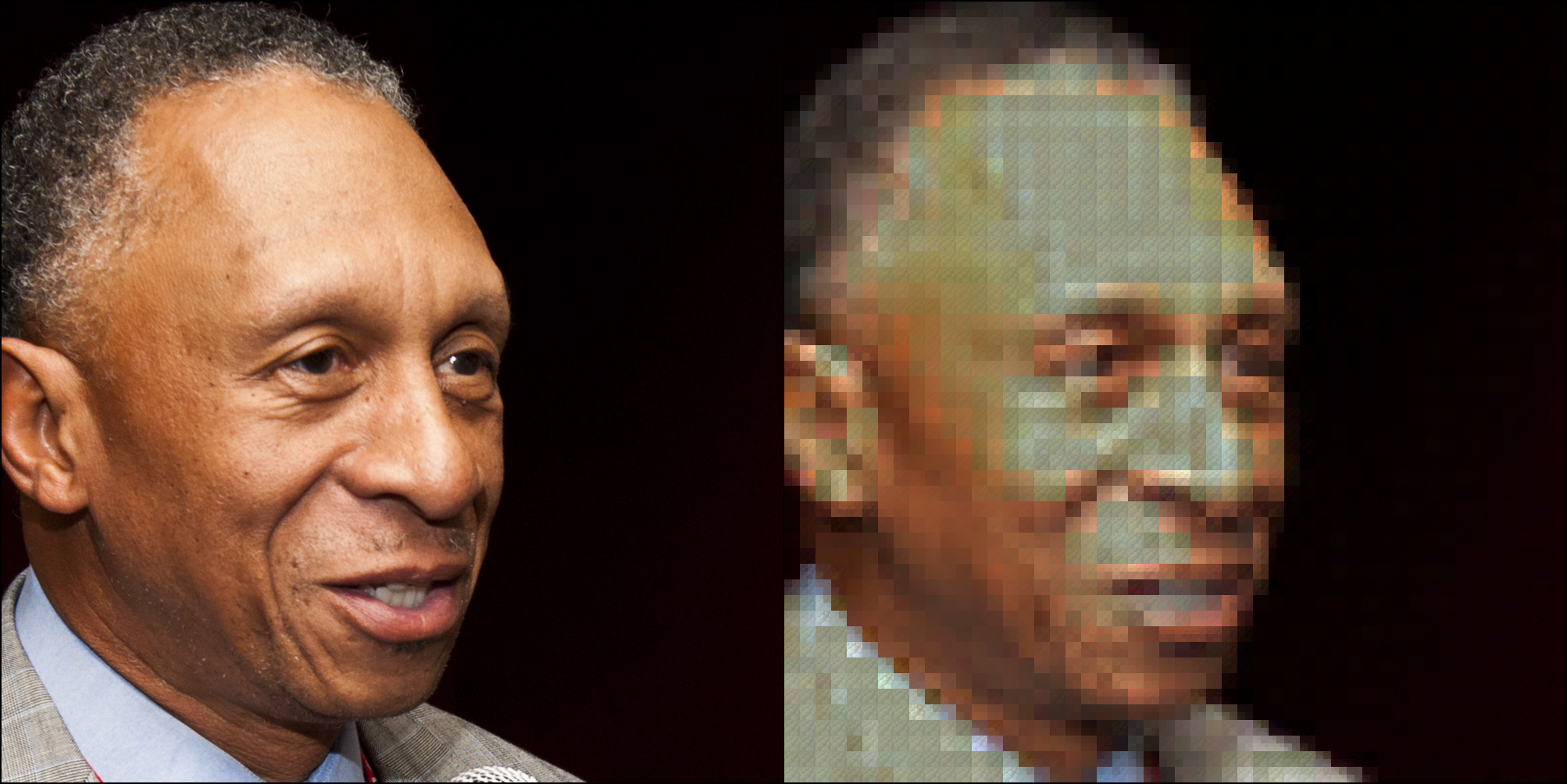} }}
\caption{Neural $\CO$ texturizer, with relative weights of $0.5$ for $\sU$ (texture source) and $1.0$ for $\sD$ (domain cells) in the $\CO$ iteration.}
\label{fig:styleae_romanesco}
\end{figure}